\documentclass{article}
\usepackage[a4paper]{geometry}

\usepackage[usenames,dvipsnames,svgnames,table]{xcolor} % added by Tomi, this should not be needed for final version
\usepackage{microtype}
\usepackage{amsmath}
\usepackage{amsfonts}
\usepackage{comment}
\usepackage{bm}
\usepackage[numbers,sort&compress]{natbib}
\usepackage{enumitem}
\usepackage{mathtools}
\usepackage{tikz}
\usepackage{subfigure}
\usetikzlibrary{arrows,decorations.pathmorphing,backgrounds,fit,positioning,shapes.symbols,chains}
\usepackage{authblk}

\usepackage{hyperref}
\ifdefined\nohyperref\else\ifdefined\hypersetup
  \definecolor{mydarkblue}{rgb}{0,0.08,0.45}
  \definecolor{mydarkred}{rgb}{0.64,0,0}
  \hypersetup{ %
    pdftitle={},
    pdfauthor={},
    pdfkeywords={},
    pdfborder=0 0 0,
    pdfpagemode=UseNone,
    colorlinks=true,
    linkcolor=mydarkblue,
    citecolor=mydarkred,
    filecolor=cyan,
    urlcolor=magenta,
    pdfview=FitH}

\newcommand{\tp}{^{\top}}

\DeclareMathOperator*{\argmax}{arg\,max}
\DeclareMathOperator{\KLt}{KL}
\DeclarePairedDelimiterX{\infdivx}[2]{[}{]}{%
  #1\;\delimsize\|\;#2%
}
\newcommand{\KL}{\KLt\infdivx}

\DeclareMathOperator{\bernoullipdf}{Bernoulli}
\DeclareMathOperator{\normalpdf}{N}
\DeclareMathOperator{\gammapdf}{Gamma}

\DeclareMathOperator{\E}{E}

\DeclareMathOperator{\logit}{logit}
\newcommand{\bw}{\bm{w}}
\newcommand{\bX}{\bm{X}}
\newcommand{\by}{\bm{y}}
\newcommand{\bgamma}{\bm{\gamma}}
\newcommand{\fgamma}{\bm{f_{\gamma}}}
\newcommand{\fw}{\bm{f_{\bw}}}
\newcommand{\bD}{\mathcal{D}}
\newcommand{\ttn}{\tilde{t}_{\normalpdf}}
\newcommand{\ttbernoulli}{\tilde{t}_{\bernoullipdf}}

\newcommand{\ttgamma}{\tilde{t}_{\gammapdf}}
\newcommand{\tmu}{\tilde{\mu}}
\newcommand{\ttau}{\tilde{\tau}}
\newcommand{\trho}{\tilde{\rho}}

\DeclareMathOperator{\eye}{\textbf{I}}
\DeclareMathOperator{\diag}{diag}

\usepackage{todonotes}
\newcommand{\todom}[1]{\todo[color=green!20, inline]{#1}}

\newcommand{\todop}[1]{\todo[color=yellow!20, inline]{#1}}

\title{Knowledge Elicitation via Sequential Probabilistic Inference for High-Dimensional Prediction \footnote{This is the pre-print version. The paper is published in Machine Learning journal. Definitive version DOI: 10.1007/s10994-017-5651-7. Link: http://rdcu.be/t9KF.}}

\author{Pedram Daee$^\dagger$} 
\author{Tomi Peltola$^\dagger$}
\author{Marta Soare$^\dagger$}  
\author{Samuel Kaski}

\affil{ Helsinki Institute for Information Technology HIIT, \\ Department of Computer Science, Aalto University \\   \texttt{firstname.lastname@aalto.fi}\\\smallskip  {\small  $^\dagger$Authors contributed equally.} }
\date{}
\begin{document}

\maketitle

\begin{abstract}
Prediction in a small-sized sample with a large number of covariates, the ``small $n$, large $p$'' problem, is challenging. This setting is encountered in multiple applications, such as precision medicine, where obtaining additional samples can be extremely costly or even impossible, and extensive research effort has recently been dedicated to finding principled solutions for accurate prediction. However, a valuable source of additional information, domain experts, has not yet been efficiently exploited. We formulate knowledge elicitation generally as a probabilistic inference process, where expert knowledge is sequentially queried to improve predictions. In the specific case of sparse linear regression, where we assume the expert has knowledge about the values of the regression coefficients or about the relevance of the features, we propose an algorithm and computational approximation for fast and efficient interaction, which sequentially identifies the most informative features on which to query expert knowledge. Evaluations of our method in experiments with simulated and real users show improved prediction accuracy already with a small effort from the expert.
\end{abstract}

%###########################################
\section{INTRODUCTION}
%###########################################

%Motivation: need for source of extra information in small n large p
%Sami's story (notes from meeting on 12 Sep): We are considering the regression problem with small n large p -> the first immediate solution is the sparse modeling -> it is still hard to achieve good results -> we bring more information to the problem -> How? this paper is about how we can get information from a user with limited budget. -> then add the motivating from Tomi's presentation 

% many times lage p small n
% additional source of info in expert users knowledge
% but how to use them jointly, allow to learn from both at the same time, integrate the user interactively in the learning process
% we propose this framework and an efficient algorithm for the case of sparse linear regression case
% validate it empirically 

Datasets with a small number of samples $n$ and a large number of variables $p$ are nowadays common. Statistical learning, for example regression, in these kinds of problems is ill-posed, and it is known that statistical methods have limits in how low in sample size they can go \citep{Donoho2009observed}. A lot of recent research in statistical methodology has focused on finding different kinds of solutions via well-motivated trade-offs in model flexibility and bias. These include strong assumptions about the model family, such as linearity, low rank, sparsity, meta-analysis and transfer learning from related datasets, efficient collection of new data via active learning, and, less prominently, prior elicitation.

There is, however, a certain disconnect between the development of state-of-the-art statistical methods and their application in challenging data analysis problems. Many applications have significant amounts of previous knowledge to incorporate into the analysis, but this is often unstructured and tacit. %, and building it into the analysis would require burdensome work from both experts in statistical methods and experts in the problem domain to tailor the model and elicit the knowledge in a suitable format for the analysis. 
Building it into the analysis would require tailoring the model and eliciting the knowledge in a suitable format for the analysis, which would be burdensome for both experts in statistical methods and experts in the problem domain.
More commonly, new methods are developed to work well in some broad class of problems and data, and domain experts use default approaches and apply their previous knowledge post-hoc for interpretation and discussion. Even when experts in both fields are directly collaborating, the feedback loop between the method development and application is often slow.

We propose to directly integrate the user into the modelling loop by formulating knowledge elicitation as a probabilistic inference process. We study a specific case of sparse linear regression with the aim of solving prediction problems where the number of available samples (``training data'') is insufficient for statistically accurate prediction. A core characteristic of the formulation is that it adapts to the feedback obtained from the expert and it sequentially integrates every piece of information before deciding on the next query for the expert. The sequential aspect of the approach efficiently reduces the burden on the expert, since the most informative queries will be asked first, thus reducing the overall number of needed interactions and allowing knowledge elicitation for high-dimensional parameters (such as the regression weights). By interactively eliciting and incorporating expert knowledge, our approach fits into the interactive learning literature. Our focus is here on the probabilistic modelling part of the interaction and we leave the design of supporting user interfaces for future work.

\subsection*{Contributions and Outline}

The outline of the paper and our main contributions are as follows. After discussing related work (Sect.~\ref{sec:related_work}), we rigorously formulate the expert knowledge elicitation as a probabilistic inference process (Sect.~\ref{sec:framework}). We study a specific case of sparse linear regression, and in particular, consider cases where the user has knowledge about the values of the regression coefficients and about the relevance of the features (Sect.~\ref{sec:sparse_linear_regression}). We present an algorithm for efficient interactive sequential knowledge elicitation for high-dimensional models that makes knowledge elicitation in ``small $n$, large $p$'' problems feasible (Sect.~\ref{sec:algo}). We describe an efficient computational approach using deterministic posterior approximations allowing real-time interaction for the sparse linear regression case (Sect.~\ref{sec:computation}). Simulation studies are presented to demonstrate the performance and to gain insight into the behaviour of the approach (Sect.~\ref{sec:experiments}). Finally, we demonstrate that real users are able to improve the predictive performance of sparse linear regression in a proof-of-concept experiment (Sect.~\ref{sec:expes_real_user}).

%###########################################
\section{RELATED WORK}\label{sec:related_work}
%###########################################

The problem we study relates to several topics studied in the literature, either by the method, goal, or by the considered setting. In this section, we highlight the main connections. 

\textbf{Interactive Learning.} 
Interactive machine learning includes a variety of ways to employ user's knowledge, preferences, and human cognition to enhance statistical learning \citep{Amershi2012, Porter2013interactive}. These methods have been used successfully in several applications, such as learning user intent~\cite{Ruotsalo2013} and preferential clustering. For instance, the semi-supervised clustering method in~\cite{must-link-cannot-link,Balcan2008} %~\todomout{In this paper, they say: "The major contribution of this paper is to give a  probabilistic framework,  in  which classification preference,  like  pairwise  relations,  can  also  be  treated  as  observations."}
uses feedback on pairs of items that should or should not be in the same cluster, to learn user preferences. In addition to the differences coming from the learning task, one notable contrast between these works and our method is that their aim is to identify user preferences or opinions, whereas our goal is to use expert knowledge as an additional source of information for an improved prediction model, by integrating it with the knowledge coming from the (small $n$) data. As a probabilistic approach, our work relates to~\cite{Cano2011a} and~\cite{House2015bayesian}, where expert feedback is used for improved learning of Bayesian networks and for visual data exploration, respectively. In Sect.~\ref{sec:framework_examples}, we show how these works can be seen as instances of the general approach we propose.
%\todotout{Need to describe carefully how our method and contributions differ from \citep{Cano2011a} and \citep{House2015bayesian}. Obvious differences are the task, model, and experiments (but the framework is very similar although they do not describe it as general as we go here). How they fit into the framework is described in Section 2.1 and doesn't need to be done in detail here.}
%% Might want to also include the references below:
%in~\cite{Paba's_journal_club}, experts can disprove the clustering returned by an algorithm, which then proposes a different clustering, until the user validates the result. A more sophisticated, interactive clustering method --> It is a workshop paper, not sure if needed to cite it yet.
%Interactive machine learning techniques have also been recently used for gaining expert knowledge to improve sparse linear regression~\cite{SoAmKa2016}. 

\textbf{Active Learning.} 
The method we propose for efficiently using expert feedback is related to active learning techniques (for a survey, see, for instance, \cite{settles2010active}), where the algorithms actively select the most informative data points to be used in prediction tasks. Our method similarly queries the user for information with the goal of maximising the information gain from each feedback and thus learning more accurate models with less feedback. The same definition of efficiency with respect to the use of samples, also connects our work with experimental design techniques, recently used for linear settings by Seeger~\cite{seeger2008bayesian}, Hern{\'a}ndez-Lobato et al. ~\cite{hernandez2013generalized}, and Ravi et al.~\cite{ravi2016design}. Our task, however, is different as we do not aim at collecting new data samples, but the additional information comes from a different source, the expert, with its respective bias and uncertainty. Indeed, our method will be most useful in cases where obtaining additional input samples would be too expensive.
%%%%%%%%%%%%%%%%%%%%%
\begin{comment}
\paragraph{(Bayesian) experimental design}
\todom{Not really sure what to include here, Maybe just mention Seeger's work? (and related - see who cited that paper?).}
\todop{ We should add both Seeger and "Generalized Spike-and-Slab Priors for Bayesian Group Feature
Selection Using Expectation Propagation" paper here.}
% UPDATE: I (Pedram) added a reference to "Generalized Spike-and-Slab..." in the active learning section.
% I don't think we necessarily need to discuss experimental design here unless there's work that is more related to our problem than Seeger's. These articles will be referred to in the query alg./computation sections.
\end{comment}
%%%%%%%%%%%%%%%%%%%

\textbf{Prior Elicitation and Privileged Information.}
Many works have studied approaches for efficient elicitation of human and, in particular, expert knowledge. In prior elicitation~\cite{OHagan06}, the goal is to use expert knowledge to construct a prior distribution for Bayesian data analysis and restrict the range of parameters to be later used in learning models. Notably, an important line of work \cite{garthwaite1988quantifying, kadane1980interactive} studies methods of quantifying subjective opinion about the coefficients of linear regression models through the assessment of credible intervals. Our approach goes beyond pure prior elicitation as the training data is used to facilitate efficient user interaction. 
Another line of work considers expert feedback as privileged information~\cite{VapnikLUPI}, where additional human knowledge is allowed in the training phase only. Differently to our method, these works typically do not consider an interactive integration of the expert knowledge with the training data, and do not model the reliability of the human feedback thus received, rather, they use it as a guideline for improving the performance of learning tasks.

%Eliciting coefficients for linear regression methods has been shown to be efficient in previous related studies. An important line of work \cite{garthwaite1988quantifying, kadane1980interactive} studies methods of quantifying subjective opinion about the coefficients of linear regression models. The prior knowledge is elicited through tasks that use hypothetical data and the assessment of credible intervals. These elicitation methods are shown to obtain prior distributions that represent well the expert's opinion, but the use of expert knowledge is not explored further. In our approach, we elicit expert's opinion more directly and this also allows to focus on more specific cost functions (such as reducing the prediction error for a specific target).

%%%%%%%%%%%%%%%%%%%%%
\begin{comment}
\todom{Only workshop papers, but interesting related topics in the \href{http://smileclinic.alwaysdata.net/ijcai16workshop/}{BeyondLabeler - Human is More Than a Labeler} @IJCAI2016.}
\end{comment}
%%%%%%%%%%%%%%%%%%%

%###########################################
\section{KNOWLEDGE ELICITATION AS INTERACTIVE PROBABILISTIC MODELLING}\label{sec:framework}
%###########################################

In the following, we formulate expert knowledge elicitation as a probabilistic inference process.
% General design considerations in interactive machine learning systems are discussed in \citet{Porter2013interactive} and \citet{Amershi2012}.

%giving references to some papers that discuss interactive ML in more general settings and in wider scope (e.g., interfaces and such) while we focus on the specific probabilistic setting.

\subsection{Key Components}

Let $y$ and $x$ denote the outputs (target variables) and inputs (covariates), and $\theta$ and $\phi_y$ the model parameters. Let $f$ encode input from the user (feedback based on the user's knowledge) and $\phi_f$ be related model parameters. We identify the following key components:
\begin{enumerate}[itemsep=1.5pt]
  \item An observation model $p(y|x,\theta,\phi_y)$ for $y$. %data model
  \item A feedback model $p(f|\theta, \phi_f)$ for user's knowledge. % user model
  \item A prior model $p(\theta, \phi_y, \phi_f)$ completing the hierarchical model description.
  \item A query algorithm and user interface that facilitate gathering $f$ iteratively from the user.
  \item Update process of the model after user interaction.
\end{enumerate}
The observation model can be any appropriate probability model. It is assumed that there is some parameter $\theta$, possibly high-dimensional, that the user has knowledge about. The user's knowledge is encoded as (possibly partial) feedback $f$ that is transformed into information about $\theta$ via the feedback model. Of course, there could be a more complex hierarchy tying the observation and feedback models, and the feedback model can also be used to model more user-centric issues, such as the quality of or uncertainty in the knowledge or user's interests.

The feedback model, together with a query algorithm and a user interface, is used to facilitate an efficient interaction with the user. The term ``query algorithm'' is used here in a broad sense to describe any mechanism that is  used to intelligently guide the user's focus in providing feedback to the system. This enables considering a high-dimensional $f$ without overwhelming the user as the most useful feedbacks can be queried first. Crucially, this enables going beyond pure prior elicitation as the observed data can be used to inform the queries via the dependence of the feedback and observation models. For example, the queries can be formed as solutions to decision or experimental design tasks that maximize the expected information gain from the interaction.

Finally, as the user's feedback is modelled as additional data, Bayes theorem can be used to sequentially update the model during the interaction. For real-time interaction, this may present a challenge as computation in probabilistic models can be demanding. It is known that slow computation can impair effective interaction \citep{Fails2003interactive} and, thus, efficient computational approaches are important.

\subsection{Overall interaction scheme}

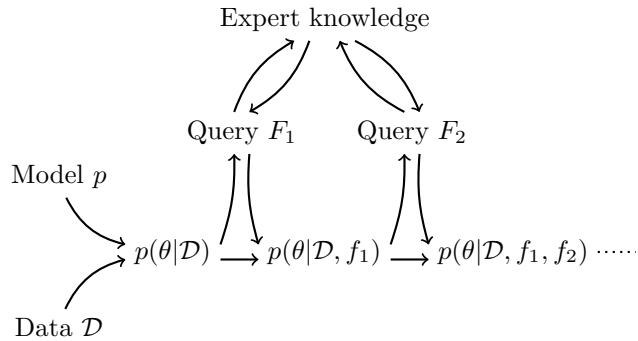
\begin{figure}[h]
  \centering
  \begin{tikzpicture}[
      %node distance=2.5cm,
      %circle
    ]
    \node[] at (-4.5,2) (DM) {Model $p$};
    \node[] at (-4.5,0) (D) {Data $\mathcal{D}$};
    \node[] at (-3,1) (p1) {$p(\theta|\mathcal{D})$};
    \node[right=0.5cm of p1] (p2) {$p(\theta|\mathcal{D},f_1)$};
    \node[right=0.5cm of p2] (p3) {$p(\theta|\mathcal{D},f_1,f_2)$};
    \node[right=0.5cm of p3] (p4) {};

    \node[above right=1 and 0.28cm of p1,anchor=south] (q1) {Query $F_1$};
    \node[above right=1 and 0.28cm of p2,anchor=south] (q2) {Query $F_2$};

    \node[above right=2.5 and 0cm of p2.north,anchor=south,rectangle] (E) {Expert knowledge};

    \path[->,bend left=-25,thick] (DM) edge ([yshift=0.1cm]p1.west);
    \path[->,bend left=25,thick] (D) edge ([yshift=-0.1cm]p1.west);

    \path[->,thick] ([yshift=-0.1cm]p1.east) edge ([yshift=-0.1cm]p2.west);
    \path[->,thick] ([yshift=-0.1cm]p2.east) edge ([yshift=-0.1cm]p3.west);
    \path[dotted,thick] (p3) edge (p4);

    \path[->,bend left=-10,thick] ([yshift=0.1cm]p1.east) edge ([xshift=-0.1cm]q1.south);
    \path[->,bend left=-10,thick] ([xshift=0.1cm]q1.south) edge ([yshift=0.1cm]p2.west);

    \path[->,bend left=-10,thick] ([yshift=0.1cm]p2.east) edge ([xshift=-0.1cm]q2.south);
    \path[->,bend left=-10,thick] ([xshift=0.1cm]q2.south) edge ([yshift=0.1cm]p3.west);

    \path[<-,bend left=-20,thick] ([xshift=0.1cm]q1.north) edge ([xshift=-0.2cm]E.south);
    \path[<-,bend left=-20,thick] ([xshift=-0.4cm]E.south) edge ([xshift=-0.1cm]q1.north);

    \path[<-,bend left=-20,thick] ([xshift=0.1cm]q2.north) edge ([xshift=0.4cm]E.south);
    \path[<-,bend left=-20,thick] ([xshift=0.2cm]E.south) edge ([xshift=-0.1cm]q2.north);
  \end{tikzpicture}
 \caption{Information flow. The parameters $\phi_y$ and $\phi_f$ are omitted from the posterior distributions for brevity.}\label{fig:infoflow}
\end{figure}

Figure~\ref{fig:infoflow} depicts the information flow. First, the posterior distribution given the observations $\mathcal{D} = \{(y_i, x_i) : i = 1,\ldots,n \}$ is computed. Then, the user is queried iteratively for feedback via the user interface and the query algorithm. The feedback is used to sequentially update the posterior distribution. The query algorithm has access to the latest beliefs about the model parameters and the predicted user behaviour, that is, the posterior predictive distribution of $f$, $p(f_{t+1}| \mathcal{D}, f_1,\ldots,f_t)$ where $f_j$ are possibly partial observations of $f$, to formulate queries and highlight the most informative interactions in the user interface.

\subsection{Examples}\label{sec:framework_examples}

The goal in this paper is to use the interaction scheme to help solve prediction problems in the ``small $n$, large $p$'' setting. The approach as described above is, however, more general and applicable to other problems. We briefly describe two earlier works that can be seen as instances of it.

\citet{Cano2011a} present a method for integrating expert knowledge into learning of Bayesian networks. The observation model is a multinomial Bayesian network with Dirichlet priors. The user provides answers to queries about the presence or absence of edges in the graph and the feedback model assumes the answers to be correct with some probability. Which edge to query about next is selected by maximising the information gain with regard to the inclusion probability of the edges. Monte Carlo algorithms are used for the computation.

\citet{House2015bayesian} present a framework for interactive visual data exploration. They describe two observation models, principal component analysis and multidimensional scaling, that are used for dimensionality reduction to visualise the observations in a two dimensional plot. They do not have a query algorithm, but their user interface allows moving points in a low-dimensional plot closer or further apart, which is interpreted by a feedback model that transforms the feedback into appropriate changes in the shared parameters with the observation model to allow exploration of different aspects of the data. Their model affords closed form updates.
%\todomout{We could (should?) also include the semi-supervised clustering paper~\cite{must-link-cannot-link} as example here, see related work (Sect. \ref{sec:related_work}). BTW, check redundancies between these two sections.}

%###########################################
\section{FEEDBACK MODELS AND QUERY ALGORITHM FOR SPARSE LINEAR REGRESSION}~\label{sec:sparse_linear_regression}
%###########################################

We next introduce the knowledge elicitation approach for sparse linear regression.

\subsection{Sparse Regression Model}% Data Model

Let $\by \in \mathbb{R}^n$ be the observed output values and $\bX \in \mathbb{R}^{n \times m}$ the matrix of covariate values. The regression is modelled with Gaussian observation model and a spike-and-slab sparsity-inducing prior \citep{george1993variable} on the regression coefficients $\bw \in \mathbb{R}^m$, and a Gamma prior on the inverse of the residual noise variance $\sigma^2$:
\begin{align}
  \by &\sim \normalpdf(\bX \bw, \sigma^2 \eye), & \\
  \sigma^{-2} &\sim \gammapdf(\alpha_{\sigma}, \beta_{\sigma}), & \nonumber\\
  w_j &\sim \gamma_j \normalpdf(0, \psi^2) + (1 - \gamma_j) \delta_0, & j=1,\ldots,m, \nonumber\\
  \gamma_j &\sim \bernoullipdf(\rho), & j=1,\ldots,m. \nonumber %\\
  %\rho &\sim \betapdf(\alpha_{\rho}, \beta_{\rho}).& \nonumber
\end{align}

Here, the $\gamma_j$ are latent binary variables indicating inclusion or exclusion of the covariates in the regression ($\delta_0$ is a point mass at zero) and $\rho$ is the prior inclusion probability controlling the expected sparsity. The $\alpha_{\sigma}$, $\beta_{\sigma}$, $\psi^2$, and $\rho$ are assumed fixed hyperparameters.

\subsection{Feedback Models}% user models

We consider two simple and natural feedback models encoding knowledge about the individual regression coefficients:
\begin{itemize}[leftmargin=*,noitemsep]
  \item User has knowledge about the value of the coefficient ($f_{w,j} \in \mathbb{R}$): 
  \begin{equation}\label{eq:feedback_coefficient}
     f_{w,j} \sim \normalpdf(w_j, \omega^2).
  \end{equation}
  \item User has knowledge about the relevance of coefficient ($f_{\gamma,j} \in \{0,1\}$ for not-relevant, relevant):
   \begin{equation}\label{eq:feedback_relevance}
    \!f_{\gamma,j}\!\sim \gamma_j\!\bernoullipdf(\pi)\!+\!(1\!-\!\gamma_j)\!\bernoullipdf(1\!-\!\pi).
   \end{equation}
%  \item Knowledge about sign of the coefficient ($\{0,1\})$):
%  \begin{equation*}
%    f_{s,j} \sim I(w_j \geq 0) \bernoullipdf(\pi) + I(w_j < 0) \bernoullipdf(1 - \pi).
%    \end{equation*}
%  \item Knowledge about probability of relevance ($(0,1)$):
%   \begin{equation*}
%    f_{p,j} \sim \betapdf(\rho_j \kappa, (1-\rho_j) \kappa).
%    \end{equation*}
\end{itemize}

Here, $\omega^2$ and $\pi$ control the uncertainty or strength of the knowledge. In detail, $\omega^2$ is the uncertainty in the user's estimate of the coefficient, and $\pi$ is the probability that the user gives correct feedback relative to the state of the covariate inclusion indicator $\gamma_j$. %We have implemented the two models and we will study them in the experiments (Sect.~\ref{sec:experiments}). 
%{\color{red} [We could also just have the first two here.]} -->Moved to discussion

\subsection{Query Algorithm}\label{sec:algo}

Our aim is to improve prediction. Thus, the user interaction should focus on aspects of the model (here, predictive features) that would be most beneficial towards this goal. We use the query algorithm to rank the features for choosing which feature to ask feedback about next. The ranking is formulated as a Bayesian experimental design task~\citep{chaloner1995}. More specifically, the feature $j^*$ that maximizes the expected information gain is chosen next:
%
%\resizebox{1.01\columnwidth}{!}
%{
\begin{equation*}
j^*\!=\!\argmax_{j\notin \mathcal{F}}\E_{p(\tilde{f}_j|\bD)}\!\left[\!\sum_i \KL{p(\tilde{y}|\bD\!,\bm{x}_i\!,\tilde{f}_j)\!}{\!p(\tilde{y}|\bD\!,\!\bm{x}_i)}\!\right]\!,
\end{equation*}
%}
%
where $j$ indexes the features, $\mathcal{F}$ is the set of feedbacks that have already been given (to simplify notation, those are here included in $\bD$), and the summation over $i$ goes over the training dataset. The information gain is defined as the Kullback--Leibler divergence ($\KLt$) between the current posterior predictive distribution $p(\tilde{y}|\bD,\bm{x}) = \int p(\tilde{y}|\bm{x}, \bm{\theta}) p(\bm{\theta}|\bD) d\bm{\theta}$, where $\bm{\theta} = (\bw, \bm{\gamma}, \sigma^2)$, and the posterior predictive distribution with the new feedback $f_j$, $p(\tilde{y}|\bD,\bm{x},f_j)$. The bigger the information gain, the bigger impact the new feedback has on the predictive distribution. Since the feedback itself will only be observed after querying the user, we take the expectation over the posterior predictive distribution of the feedback $p(\tilde{f}_j|\bD)$. More details about the Bayesian experimental design are provided in the supplementary material (Sec.~\ref{sec:Bayes_ed}).  

%where $j$ indexes the features, $\mathcal{F}$ is the set of feedbacks that have already been given (to simplify notation, those are here included in $\bD$), and the summation over $i$ goes over the training dataset. More details about the Bayesian experimental design are provided in the supplementary material (Sec.~\ref{sec:Bayes_ed}). 

We note that, were the predictive distribution of $y$ Gaussian, the problem would be simple. The expected information gain would be independent of $y$ and the actual values of the feedbacks (when feedback is on values of the regression coefficients) and would only depend on the $x$ and which features were given feedback on \citep{seeger2008bayesian}. The sparsity-promoting prior, however, makes the problem non-trivial.

\subsection{Computation}\label{sec:computation}

The model does not have a closed form posterior distribution, predictive distribution, or solution to the information gain maximization problem. To achieve fast computation, we use deterministic posterior approximations. Expectation propagation \citep{minka2001expectation} is used to approximate the spike-and-slab prior \citep{Lobato2015ML} and the feedback models, and variational Bayes (e.g., \citep[Chapter 10]{Bishop2006}) is used to approximate the residual variance $\sigma^2$. The form of the posterior approximation for the regression coefficients $\bw$ is Gaussian. The posterior predictive distribution for $y$ is also approximated as Gaussian. Details are provided in the supplementary material~(Sect.~\ref{sec:posterior_approximation}).

Expectation propagation has been found to provide good estimates of uncertainty, which is important in experimental design \citep{seeger2008bayesian, hernandez2013generalized, Lobato2015ML}. In evaluating the expected information gain for a large number of candidate features, running the approximation iterations to full convergence for each, however, is too slow. We follow the approach of \citet{seeger2008bayesian, hernandez2013generalized} in computing only a single iteration of updates on the essential parameters for each candidate. We show in the results that this already provides a good performance for the query algorithm in comparison to random queries. Details on the computations are provided in the supplementary material~(Sect.~\ref{sec:single_iteration_update}).

%###########################################
\section{EXPERIMENTS}\label{sec:experiments}
%###########################################

The performance of the proposed method (Sect. \ref{sec:sparse_linear_regression}) is evaluated in several ``small $n$, large $p$'' regression problems on both simulated and real data. A proof-of-concept user study is presented to demonstrate the feasibility of the method with real users. \footnote{All codes and data are available in \href{https://github.com/HIIT/knowledge-elicitation-for-linear-regression}{https://github.com/HIIT/knowledge-elicitation-for-linear-regression}.} %\footnote{All codes and data are available in the supplementary material. A link for public access will be included in the published version.}

\subsection{Simulated Data}\label{sec:heatmap_dimension}
We use synthetic data to study the behaviour of the approach in a wide range of controlled settings. % for a wide range of values for the considered parameters. 

\textbf{Setting.}
The covariates of $n$ training data points are generated from  $\bX \sim \normalpdf(\bm{0}, \eye) $. Out of the $m$ regression coefficients 
%$\bw \in \mathbb{R}^m$, 
$w_1,\dots,w_m \in \mathbb{R}$, $m^*$ are generated from $w_j \sim \normalpdf(0, \psi^2)$ and the rest are set to zero. 
%considered as zero. 
The observed output values are generated from $\by \sim \normalpdf(\bX \bw, \sigma^2 \eye)$. We consider cases where the user has knowledge about the value of the coefficients (Eq.~\ref{eq:feedback_coefficient} with noise value $\omega=0.1$) and where the user has knowledge about non-relevant/relevant features (Eq.~\ref{eq:feedback_relevance} with $\gamma_j=1$ if $w_j$ is non-zero, and $\gamma_j=0$ otherwise, and $\pi=0.95$). 
For a generated set of training data, all algorithms query feedback about one feature at a time. Mean squared error (MSE) is used as the performance measure to compare the query algorithms. For the simulated data setting, we use the known data-generating values for the fixed hyperparameters, namely: $\psi^2=1,~\rho=m^*/m, ~\text{and}~\sigma^2=1$ (here we do not use the distribution assumption on $\sigma^2$).

\begin{figure*}
\centering
    \subfigure[Feedback on coefficients' values]{\label{fig:sim_study_all_weights_dims}
    \includegraphics[width=0.45\textwidth,keepaspectratio]{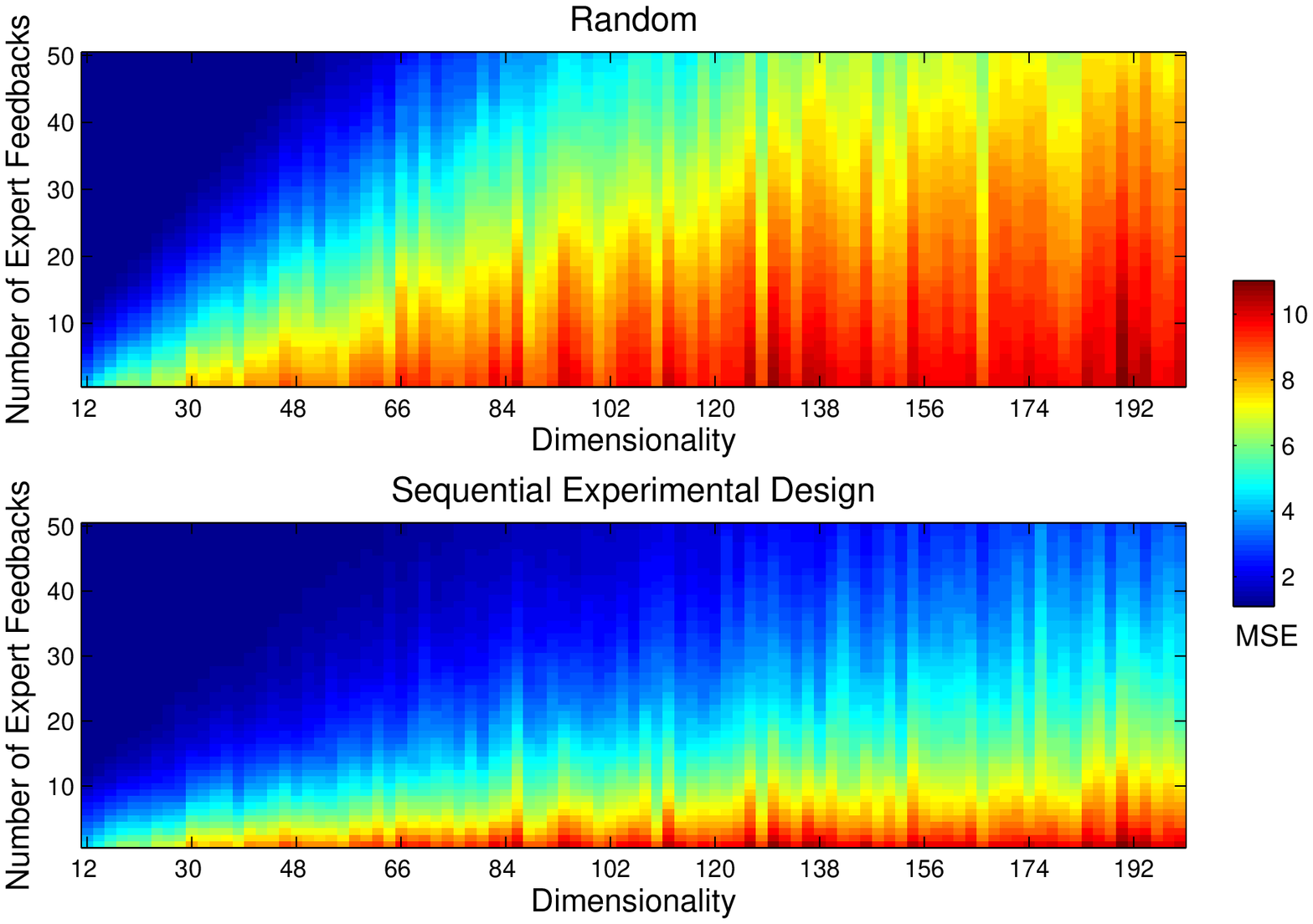}
    }
    \subfigure[Feedback on coefficients' relevances]{\label{fig:sim_study_all_relevance_dims}
    \includegraphics[width=0.45\textwidth,keepaspectratio]{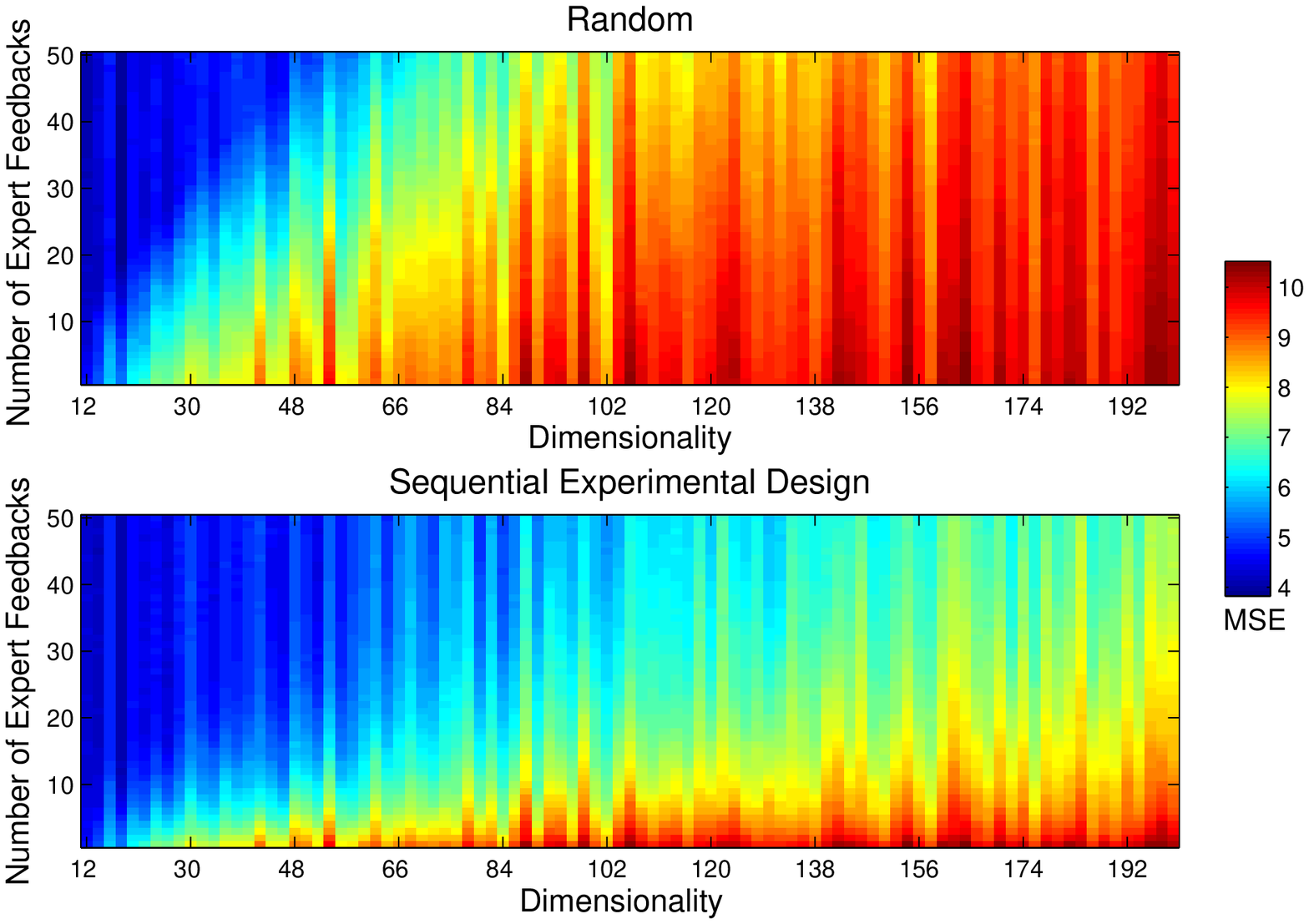}
    }
\caption{Mean squared errors in simulated settings with increasing dimensionality. The number of relevant coefficients $m^*=10$ and the number of training data points $n=10$. The MSE values are averages over 100 independent runs.}
\label{fig:heatmap_dimensions}
\end{figure*} 

\textbf{Results.}  In Fig.~\ref{fig:heatmap_dimensions}, we consider a ``small $n$, large $p$'' scenario, with $n=10, m^*=10$ and with increasing dimensionality (hence also increasing sparsity) from $m=12,\dots,200$. The heatmaps show the average MSE values over 100 runs (repetitions of the data generation) for both feedback models, as obtained by our sequential experimental design algorithm and by a strategy that randomly selects the sequence of features on which to ask for expert feedback. The result shows that our method achieves a faster improvement in the prediction, starting from the very first user feedbacks, for both feedback types, and at all the dimensionalities. 

Notably, in the case of the random strategy, the performance decreases rapidly with the growing dimensionality (even with 50 feedbacks, in the setting with 200 dimensions, the prediction error for random strategy stays high), while the user feedback via the sequential experimental design is informative enough to provide good predictions even in large dimensionalities. Comparing the two types of feedback, the feedback on the coefficient values gives better performance for both strategies.

Sect.~\ref{sec:heatmap_traindata} in the supplement shows heatmaps for the same setting but with a fixed dimension $m=100$ and a varying number of training data $n=5,\dots,50$. For those experiments, we can again see superior improvement for the sequential experimental design compared to random, for both feedback models, and in particular for small sample sizes. 
Moreover, a comparison of the sequential experimental design algorithm to its non-sequential version (Sect.~\ref{sec:seq_vs_nonseq} in the supplement) shows that the former achieves a better performance, indicating that the user feedback affects the next query. 
Finally, for further insight into the behaviour of the approach, a simulation experiment with $n=10$ in Sect.~\ref{sec:trainingerror} in the supplementary material shows that the training set error begins to increase as a function of the number of feedbacks while the test error decreases. This happens because the initial fit exhausts the information in the training data, but at this small sample size is insufficient to provide good generalization performance.

\subsection{Review Rating Prediction}

We test our method for the task of predicting review ratings from textual reviews in subsets of Amazon and Yelp datasets. Each review is one data point, and each distinct word is a feature with the corresponding covariate value given by the number of appearances of the word in the review. In addition to being fit for sparse linear regression models (as shown in previous studies, for instance, in~\cite{Lobato2015ML}), we also chose this type of dataset due to the uncomplicated interpretation of the features, which allows us to easily test our method on real users. %, without putting a constraint on their level of domain expertise required in order to receive reliable feedback. 
%
%Indeed, we consider a small number of samples in the training set, and use the rest of the datasets for constructing (pseudo) ground-truth knowledge for simulated users (Sect.~\ref{sec:simulated_users}), respectively, 10 real user feedbacks (Sect.~\ref{sec:expes_real_user}). 

%We evaluate the performance of the algorithms by studying the efficiency of the sequence of queries they suggest (that is, their sequential reduction of the loss). %Before reporting and commenting the results, we briefly describe the datasets. 

\subsubsection{Datasets}

\paragraph{Amazon data.}
% source
The Amazon data is a subset of the sentiment dataset of~\cite{Blitzer2007}. 
% description
This dataset\footnote{\url{https://www.cs.jhu.edu/~mdredze/datasets/sentiment/}} contains textual reviews and their corresponding 1-5 star ratings for Amazon products. Here, we only consider the reviews for products in the \textit{kitchen appliances} category, which amounts to 5149 reviews. The preprocessing of the data follows the method described in~\cite{Lobato2015ML}, where this dataset was used for testing the performance of a sparse linear regression model.
% preprocessing
Each review is represented as a vector of features, where the features correspond to unigrams and bigrams, as given by the data provided by~\cite{Blitzer2007}. For each distinct feature and for each review, we created a matrix of occurrences and only kept for our analysis the features that appeared in at least 100 reviews, that is, 824 features.

\paragraph{Yelp data.} The second dataset we use is a subset of the YELP (academic) dataset\footnote{\url{https://www.yelp.com/dataset_challenge}}. The dataset contains 2.7 million restaurant reviews with ratings ranging from 1 to 5 stars (rounded to half-stars). %The textual reviews from costumers were collected from 2004 to 2016. 
Here, we consider the 4086 reviews from the year 2004. Similarly to the preprocessing done for Amazon data, each review is represented as a vector of features (distinct words). After removing non-alphanumeric characters from the words and removing words that appear fewer than 100 times, we have 465 words for our analysis.
\begin{table}[h]
\centering
    \begin{tabular}{c|c|c}
    Dataset Subset & Reviews & Features \\
    \hline
    Yelp &  4086 & 465 %(100 app) 828 (50 app)
    \\
    Amazon & 5159  & 824
    \end{tabular}
%\vspace{-0.05cm}    
\caption{Sizes of the datasets used in experiments.}
\end{table}

\subsection{Simulated User Feedback}\label{sec:simulated_users}

For all experiments on Amazon and Yelp datasets, we proceeded as follows: First, each dataset was partitioned in three parts: (1) a training set of 100 randomly selected reviews, (2) a test set of 1000 randomly selected reviews, and (3) the rest as a ``user-data set'' for constructing simulated user knowledge. The data were normalised to have zero mean and unit standard deviation on the training and user-data sets. The simulated user feedback was generated based on the posterior inclusion probabilities $\mathbf{E}[\gamma]$ in a spike-and-slab model trained on the user-data partition. We only considered the more realistic case where the user can give feedback about the relevance of the words. For a word $j$ selected by the algorithm, the user gives feedback that the word is \textit{relevant} if $\mathbf{E}[\gamma_j]>\pi$, \textit{not-relevant} if $\mathbf{E}[\gamma_j]<1-\pi$, and \textit{uncertain} otherwise. The intuition is that if the user-data indicate that a feature is zero/non-zero with high probability, then the simulated user would select that feature as \textit{not-relevant}/\textit{relevant}. However, for \textit{uncertain} words, the feedback iteration passes without receiving any feedback. The model parameters were set to $\pi=0.9$, $\psi^2=0.01$, $\alpha_{\sigma}=1$, $\beta_{\sigma}=1$, and $\rho=0.3$.

\subsubsection{Results} 

We compare three query algorithms:
\begin{itemize}[leftmargin=*,noitemsep,topsep=0pt]
 \item random feature suggestion \textit{(green line, triangle up)},
 \item an strategy that knows the relevant features beforehand (inferred from the posterior inclusion probabilities over all data) and asks exclusively about them first, and then chooses at random from the features not already selected \textit{(red line, triangle down)}\footnote{Although unrealistic, this ``oracle'' strategy allows to see the performance gain obtainable by an intuitively good strategy which first queries experts about the relevant features.},
 %The reason for including this unrealistic method in the results is to have an ``oracle'' lower-bound on the minimal prediction error attainable by asking feedback on relevant features only, at the beginning of the user interaction.},
 \item our sequential experimental design algorithm (Sect.~\ref{sec:algo}) \textit{(blue line, squares)}.
% \item the non-sequential version of our algorithm, which chooses the sequence of features to be queried before observing user feedback \textit{(magenta line, circles)}.
\end{itemize} 
\begin{figure}[ht]
%\vspace{-0.3cm}
\centering
   \subfigure[Amazon data]{\label{fig:sim_user_Amazon}
       \includegraphics[width=0.44\columnwidth,keepaspectratio]{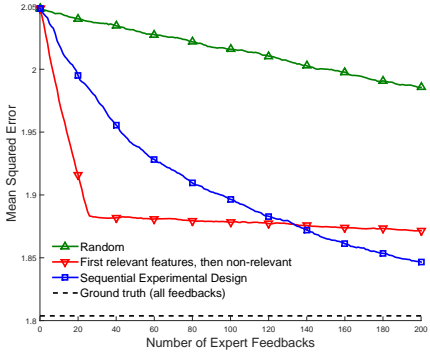}
   }
   \subfigure[Yelp data]{\label{fig:sim_user_Yelp}
       \includegraphics[width=0.44\columnwidth,keepaspectratio]{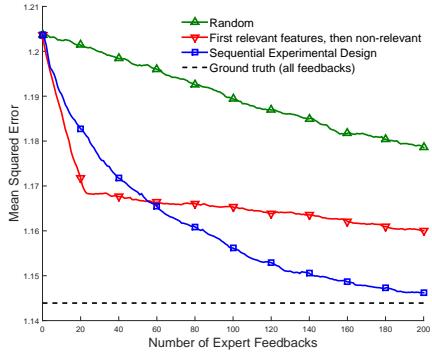}
   }
\caption{Mean squared errors when user feedback is on relevance of features for Amazon %(Fig.~\ref{fig:sim_user_Amazon}), 
and Yelp %(Fig.~\ref{fig:sim_user_Yelp}) 
data. The MSE values are averages over 100 independent runs.}
\label{fig:sim_user_amazon_yelp}
\end{figure}
All algorithms query feedback about one feature at a time and MSE is used as the performance measure. The ground truth line represents the MSE after receiving user feedback for all words in each dataset.

A first observation is that the use of additional knowledge coming from the simulated expert indeed reduces the prediction errors, for all algorithms and on both datasets. Yet, the reduction in the prediction error differs significantly depending on whether the methods manage to query feedback on the most informative features first. Indeed, the goal is to make the elicitation as little burdensome as possible for the experts. To reach the goal, a strategy needs to rapidly extract a maximal amount of information from the expert, which here amounts to the careful selection of the features on which to query feedback. As expected, the random query selection strategy has a constant and slow improvement rate, as the number of feedbacks grows, leaving a big gap from the ground-truth performance in both datasets, even after 200 user feedbacks. In contrast, the (unrealistic) strategy that first asks about relevant features begins with a steep increase in performance for the first iterations (only 26 words for Amazon and 23 for Yelp are marked as relevant, as computed from the full dataset), then it continues with a very slow improvement rate coming from asking  non-relevant words. Our method manages to identify the informative features rapidly and thus has a higher improvement compared to random from the first user feedbacks. In the case of Yelp data, our strategy manages to be very close to the strategy knowing the relevant words in the initial feedbacks and then getting very close to the ground-truth after 200 interactions. Furthermore, there is a significant gap compared to the random strategy for all amounts of feedbacks. In the more difficult (in terms of rating prediction error and size of dimensions) Amazon dataset, although the gap to the random strategy is clear, our strategy exceeds the level of information obtained in the 26 non-zero features only after 140 feedbacks.

\subsubsection{Expert Knowledge Elicitation vs.\ Collecting More Samples}
%extra feedback vs extra data point - given the "quality" of user feedback and "quality" of datapoint (quality =?? how reliable/informative the extra info is)
We next contrast the improvements in the predictions brought by eliciting the expert feedback to improvements gained by adding samples from the user-data set to the training set. For the latter, we use two alternative strategies: randomly selecting a sequence of reviews to be included in the training set, and an active learning strategy, which selects samples based on maximizing expected information gain (an adaptation of the method in \cite{seeger2008bayesian}).
%For the rating prediction setting presented above, we can easily consider training data with increasing samples and compare the gain in prediction improvement brought by adding samples, versus the gain obtained with increasing number of user feedback. 

\begin{comment}
%the following table is the submitted version
\begin{table}[h!]
\centering
\resizebox{\columnwidth}{!}{  
    \begin{tabular}{c|c|c|c|c|c}
     & \multicolumn{2}{c|}{More Samples}  & \multicolumn{2}{c|}{More Feedback} \\
    \hline
   \textsc{MSE} & \textsc{Random} & \textsc{Active}~\cite{seeger2008bayesian} & \textsc{Random} & \textsc{SeqExpDes}  \\
    \hline
    1.21  & 24 & 3  & 47 & 5
    \\    
    1.20  & 48 & 7  & 108 & 13
    \\
    1.19 & 103 & 12 & 175 & 25
    \\
    1.18 & 167 & 21 & 247 & 41
    \\
    1.17 & 226 & 35 & 320 & 73
%    \\
%    l.16 & 251  & 67 & 397 & 122 
    \end{tabular}
}
\caption{Number of samples/feedbacks needed to reach a particular MSE level in Yelp dataset. The values are averages over 100 independent runs.}
\label{table:realdata_simuser_yelp}
\end{table}
\end{comment}

%the following table is based on the new results
\begin{table}[h!]
\centering
%\resizebox{\columnwidth}{!}{  
    \begin{tabular}{c|c|c|c|c|c}
     & \multicolumn{2}{c|}{More Samples}  & \multicolumn{2}{c|}{More Feedback} \\
    \hline
   \textsc{MSE} & \textsc{Random} & \textsc{Active}~\cite{seeger2008bayesian} & \textsc{Random} & \textsc{SeqExpDes}  \\
    \hline
    
    1.20  & 21 & 3  & 30 & 3
    \\
    1.19 & 55 & 6 & 96 & 11
    \\
    1.18 & 94 & 12 & 185 & 25
    \\
    1.17 & 146 & 22 & 266 & 46
    \\
    l.16 & 241  & 44 & 324 & 85 
    \end{tabular}
%}
\caption{Number of samples/feedbacks needed to reach a particular MSE level in Yelp dataset. The values are averages over 100 independent runs.}
\label{table:realdata_simuser_yelp}
\end{table}

% How to read the table?
Table~\ref{table:realdata_simuser_yelp} shows how many \textit{feedbacks} (for the knowledge elicitation strategies in the last two columns: random and our method; see Sect.~\ref{sec:algo}) and respectively how many \textit{additional samples} (that is, additional reviews to be included in the train set) are needed to reach \textit{set levels of MSE}, noting that all strategies have the same ``small $n$, large $p$'' regression setting as a starting point, with $n=100$ and a corresponding MSE of 1.2036. %The strategies considered for eliciting expert feedback are the random selection of the features to be queried (column 4), and our sequential experimental design method (column 5), as presented in Sect.~\ref{sec:algo}. 
 %The table allows to make an interesting comparison between the information from these two sources. Thus, 
 
Even with the relatively weak type of expert feedback (feedback on the relevance of features), a specific performance is reached by a comparable number of expert feedbacks and additional data. For instance, the same level of MSE=1.18 is obtained either by asking an expert about the relevance of 25 features and by actively selecting 12 extra samples. When the active selection is not possible, we can see that the same information gain requires 94 additional randomly selected samples. Naturally, the results obtained are specific for this Yelp data and for the feedback model we assume. Nevertheless, the comparison shows the potential of expert knowledge elicitation in prediction for settings where actively selecting samples is not possible, or even more so, when getting additional samples is impossible or very expensive. The same observations and intuitions about the information gain comparison remain valid for the Amazon data (see Sect.~\ref{sec:amazon_info_gain_comparison}).

\subsection{User Study}\label{sec:expes_real_user}%Real User Feedback

The goal of the user study is to investigate the prediction improvement and convergence speed of the proposed sequential method based on human feedback. Our focus is on testing the accuracy of feedback from real users on the easily interpretable Amazon data rather than on details of the user interface. Hence, we asked ten university students and researchers to go through all the 824 words and give us feedback in the form of \textit{not-relevant}, \textit{relevant}, or \textit{uncertain}. This allowed for a fast collection of feedbacks and we could use the pre-given feedback to test the effectiveness of several query algorithms. We assumed that the algorithms had access to 100 training data and at each iteration they could query the pre-given feedback of the participant about one word. The whole process was repeated for 40 independent runs, where training data were randomly selected. The hyperparameters of the model were set to the same values as in the simulated data study with the only difference that the strength of user knowledge was lowered to $\pi=0.7$.

\begin{figure}[h]
%\vspace{-0.3cm}
\centering
\includegraphics[width=0.6\columnwidth,keepaspectratio]{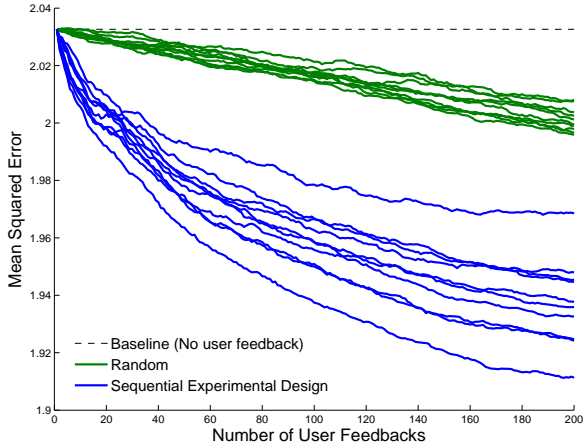}
%\vspace{-0.35cm}
\caption{Mean squared errors for ten participants. Values are averages over 40 independent runs. }
\label{fig:user_study_mse}
\end{figure}
Fig.~\ref{fig:user_study_mse} shows the average MSE improvements for each of the 10 participants, when using our proposed method and the random query order. From the very first feedbacks, the sequential experimental design approach performs better for all users and captures the expert knowledge more efficiently. The random strategy exhibits a relatively constant rate of performance improvement with the increasing number of feedbacks, while the sequential experimental design shows faster improvement rate in the beginning implying that it can query about the more important features first.

To further quantify the statistical evidence for the difference, we computed the paired-sample \textit{t} tests between the random suggestion and the proposed method at each iteration (green and blue curves in Fig.~\ref{fig:user_study_mse}). Already after the first feedback, the difference between the methods is significant at the Bonferroni corrected level $\alpha = 0.05/200$. Further analysis about the convergence speed and the suggested words are reported in the supplementary (Sect.~\ref{sec:user_study_supplementary}). 

%To further quantify the statistical evidence for the difference, we computed the paired-sample \textit{t} test between random suggestion and the proposed method for each participant at each iteration. Already after nine feedbacks, the difference between the methods is significant at the Bonferroni corrected level $\alpha = 0.05/200$. Further analysis about the convergence speed and the suggested words are reported in the supplementary (Sect.~\ref{sec:user_study_supplementary}). 

%###########################################
\section{CONCLUSION}
%###########################################

We presented a knowledge elicitation approach for high-dimensional sparse linear regression. The results for ``small $n$, large $p$'' problems in simulated and real data with simulated and real users, and with user knowledge on the regression weight values and on the relevance of features, showed improved prediction accuracy already with a small number of user interactions. The knowledge elicitation problem was formulated as a probabilistic inference process that sequentially acquires and integrates user knowledge with the training data. Compared to pure prior elicitation, the approach can facilitate richer interaction and be used in knowledge elicitation for high-dimensional parameters without overwhelming the user.

As a by-product of our study, we noticed that even for the rather weak feedback on the relevance of features, the number of expert feedbacks and the number of randomly acquired additional data samples needed to reach a certain level of MSE reduction were of the same order. Although this observation was obtained on a noisy dataset and for a simplifying user interaction setting, the fact that the considered feedback type was rather weak sets the ground for a further and more robust comparison of the performance gain obtained from these two different sources of information.

%The presented knowledge elicitation approach general, and as all assumptions have been explicated as a probabilistic model, the approach can be rigorously analyzed and changed to match specifics of other knowledge elicitation settings.
The presented knowledge elicitation method is general, and as all assumptions have been explicated as a probabilistic model, the approach can be rigorously analyzed and tailored to match specifics of other knowledge elicitation settings. The presented results considered rather simple types of feedback as a proof-of-concept of the approach. In future, we will work on extending the types of interactions and outlining new types of interactive machine learning problems.

\subsubsection*{Acknowledgements}
This work was financially supported by the Academy of Finland (Finnish Center of Excellence in Computational Inference Research COIN; grants 295503, 294238, 292334, and 284642), Re:Know funded by TEKES, and MindSee (FP7–ICT; Grant Agreement no 611570). We thank Juho Piironen for comments that  improved the article.

%\newpage

% \subsubsection*{References}

%\printbibliography

\bibliographystyle{plain} 
%\bibliography{user_interaction}	

%\defaultbibliographystyle{unsrtnat}
%\defaultbibliography{user_interaction}
%\putbib
%\end{bibunit}

\newpage
%###########################################
%###########################################
%###########################################
%###########################################
%###########################################
\appendix
%\onecolumn

%\begin{bibunit}

%###########################################  
\section{Gaussian Linear Regression with Spike and Slab Prior}
%###########################################

\subsection{Model}

The posterior distribution of the regression model is
\begin{equation*}
%  p(\bw, \sigma^2, \bgamma, \rho|\bD) \propto p(\fgamma|\bgamma) p(\fw|\bw) p(\by | \bX, \bw, \sigma^2) p(\sigma^{-2}) p(\bw | \bgamma) p(\bgamma | \rho) p(\rho),
  p(\bw, \sigma^2, \bgamma|\bD) \propto p(\fgamma|\bgamma) p(\fw|\bw) p(\by | \bX, \bw, \sigma^2) p(\sigma^{-2}) p(\bw | \bgamma) p(\bgamma),
\end{equation*}
where $\bD = (\by, \bX, \fgamma, \fw)$ are the training data observations together with the sets of observed user feedback and
\begin{align*}
  p(\fgamma|\bgamma) &= \prod_{j \in \mathcal{F}_{\gamma}} \left[\gamma_j \bernoullipdf(f_{\gamma,j}|\pi) + (1 - \gamma_j) \bernoullipdf(f_{\gamma,j} | 1 - \pi) \right], \\
        p(\fw|\bw) &= \prod_{j \in \mathcal{F}_w} \normalpdf(f_{w,j} | w_j, \omega^2), \\
  p(\by | \bX, \bw, \sigma^2) &= \normalpdf(\by | \bX \bw, \sigma^2 \eye), \\
  p(\sigma^{-2}) &= \gammapdf(\sigma^{-2}|\alpha_{\sigma}, \beta_{\sigma}), \\
  p(\bw | \bgamma) &= \prod_j \left[\gamma_j \normalpdf(w_j|0, \psi^2) + (1 - \gamma_j) \delta_0(w_j) \right], \\
  p(\bgamma) &= \prod_j \bernoullipdf(\gamma_j|\rho). %, \\
  %p(\bgamma | \rho) &= \prod_j \bernoullipdf(\gamma_j|\rho). %, \\
  %p(\rho) &= \betapdf(\rho | \alpha_{\rho}, \beta_{\rho}).
\end{align*}

Here, $\mathcal{F}_{\gamma}$ and $\mathcal{F}_w$ denote the sets of indices of the features that have received relevance feedback and weight feedback, respectively. $\pi$, $\omega^2$, $\alpha_{\sigma}$, $\beta_{\sigma}$, and $\psi^2$ are assumed fixed hyperparameters. The parametrizations of the distributions follow \citet{Gelman2013} and we use the generic $p(\cdot)$ notation, where it is understood that the parameters identify the separate terms.

\subsection{Posterior approximation}\label{sec:posterior_approximation}

The corresponding posterior approximation is
\begin{equation*}
  %q(\bw, \sigma^{-2}, \bgamma, \rho) = q(\bw) q(\sigma^{-2}) q(\bgamma) q(\rho),
  q(\bw, \sigma^{-2}, \bgamma) = q(\bw) q(\sigma^{-2}) q(\bgamma),
\end{equation*}
where, using bar to distinguish the parameters of the posterior approximation,
\begin{align*}
  q(\bw) &= \normalpdf(\bw | \bar{\bm{m}}, \bar{\bm{\Sigma}}), \\
  q(\sigma^{-2}) &= \gammapdf(\sigma^{-2} | \bar{\alpha}_{\sigma}, \bar{\beta}_{\sigma}), \\
  q(\bgamma) &= \prod_j \bernoullipdf(\gamma_j | \bar{\rho}_j), %\\
  %q(\rho) &= \betapdf(\rho | \bar{\alpha}_{\rho}, \bar{\beta}_{\rho}),
\end{align*}
and the site term approximations are
\begin{align*}
  p(\fgamma|\bgamma) &\approx \prod_{j \in \mathcal{F}_{\gamma}}\ttbernoulli(\gamma_j | \trho^{f_{\gamma}}_j), \\
        p(\fw|\bw) &= \prod_{j \in \mathcal{F}_w} \ttn(w_j | \tmu^{f_w}_j, \ttau^{f_w}_j), \\
  p(\by | \bX, \bw, \sigma^2) &\approx \ttn(\bw | \tilde{\bm{\mu}}^y, \tilde{\bm{\Gamma}}^y) \ttgamma(\sigma^{-2}|\tilde{\alpha}^y, \tilde{\beta}^y), \\
  p(\sigma^{-2}) &= \ttgamma(\sigma^{-2}|\alpha_{\sigma}-1, -\beta_{\sigma}), \\
  p(\bw | \bgamma) &\approx \prod_{j \in \mathcal{F}_{\gamma}}\ttn(\gamma_j | \tmu^w_j, \ttau^w_j) \ttbernoulli(\gamma_j|\trho^w_j), \\
  p(\bgamma) &= \prod_j \ttbernoulli(\gamma_j|\logit(\rho)), \\
  %p(\bgamma | \rho) &\approx \prod_j \ttbernoulli(\gamma_j|\trho^{\gamma}_j) \ttbeta(\rho|\talpha^{\gamma}_j,\tbeta^{\gamma}_j) \\
  %p(\rho) &= \ttbeta(\rho | \alpha_{\rho} - 1, \beta_{\rho} - 1),
\end{align*}
where $\tilde{t}_{\cdot}$ denote the exponential family forms of the corresponding distributions parametrized by the precision-adjusted mean and precision for normal distribution, and the natural parameters for Bernoulli and gamma distributions. Note that the terms $p(\sigma^{-2})$, $p(\fw|\bw)$, and $p(\bgamma)$ need not be approximated as they are already of the correct exponential family form.

The parameters of the full approximation can be identified from the products of the corresponding site term approximations and are
\begin{align*}
  \bar{\bm{m}} &= \bar{\bm{\Sigma}} (\tilde{\bm{\mu}}^y + \bm{\tmu}^w + \bm{\tmu}^{f_w}), \\
  \bar{\bm{\Sigma}} &= (\tilde{\bm{\Gamma}}^y + \diag(\bm{\ttau}^w) + \diag(\bm{\ttau}^{f_w}))^{-1}, \\
  \bar{\alpha}_{\sigma} &= \alpha_{\sigma} + \tilde{\alpha}^y, \\
  \bar{\beta}_{\sigma} &= \beta_{\sigma} - \tilde{\beta}^y, \\
  \bar{\rho}_j &= \frac{1}{1 + \exp(-(\trho^w_j + \logit(\rho) + \trho^{f_{\gamma}}_j))}, %\\
  %\bar{\alpha}_{\rho} &= \alpha_{\rho} + \sum_j \talpha^{\gamma}_j, \\
  %\bar{\beta}_{\rho} &= \beta_{\rho} + \sum_j \tbeta^{\gamma}_j,
\end{align*}
where $\diag(\cdot)$ is a diagonal matrix with the parameter as the diagonal and feedback term approximation parameters are zero for feedbacks that have not been observed.

\subsection{Computation of the posterior approximation}\label{sec:single_iteration_update}

Expectation propagation (EP) and variational Bayes (VB) inference are used to find the parameters of the posterior approximation \citep{minka2001expectation, minka2005divergence, Bishop2006}. Expectation propagation for linear regression with spike and slab prior has been introduced by \citet{hernandez2008regulator} (see \citep{Lobato2015ML} for a more extensive treatment). We update the $\ttn(\bw | \tilde{\bm{\mu}}^y, \tilde{\bm{\Gamma}}^y)$ and $\ttgamma(\sigma^{-2}|\tilde{\alpha}^y, \tilde{\beta}^y)$ term approximations using VB and all other terms using EP.

The parameter update steps in the algorithm, to be iterated until convergence, are
\begin{enumerate}
  \item $p(\bw | \bgamma)$ approximation using parallel EP update.
  %\item $p(\bgamma | \rho)$ approximation using parallel hybrid VB-EP update.
  \item $p(\by | \bX, \bw, \sigma^2)$ approximation using VB update.
  \item $p(\fgamma|\bgamma)$ approximation using parallel EP update.
\end{enumerate}

%{\color{red} [the following might not be necessary]} 
%Maybe does not hurt to have them.

The individual terms are updated following the pattern in~\citep{minka2001expectation}:
\begin{enumerate}
  \item Computation of the cavity distribution, $q^{\backslash}(\cdot) \propto \frac{q(\cdot)}{\tilde{t}(\cdot)}$.

 In the natural parametrization, this corresponds to subtracting the parameters of the site approximation from the parameters of the full approximation for the processed model parameter. 
 \item Minimization of the Kullback--Leibler divergence between the approximation $q$ and the tilted distribution, $\hat{p}(\cdot) \propto p(\cdot) q^{\backslash}(\cdot)$.

   For the EP update, $\KL{\hat{p}}{q}$ and for the VB, $\KL{q}{\hat{p}}$. The former corresponds to setting the moments of the sufficient statistics of $q$ to match those of $\hat{p}$, and the latter has solution $q(\cdot)^{new} \propto \exp(\E_{q_{-\cdot}}[\log \hat{p}(\cdot)])$, where the expectation is over the approximate posterior of all other model parameters than the one that is being processed \citep{minka2005divergence, Bishop2006}.
 \item Updating of the parameters of the site approximation, $\tilde{t}^{new} \propto \frac{q(\cdot)^{new}}{q^{\backslash}(\cdot)}$.
   
   This can be thought of as an inverse of the step 1 to now get the updated site approximation and, in the natural parametrization, is a subtraction of the cavity parameters from the parameters of the new full approximation. We use damping of the updates (the parameters are set to a convex combination of the old parameters and the new parameters computed above) \citep{minka2002expectation}.
\end{enumerate}

All of the computation have closed form solutions. %The equations for the EP update of spike and slab prior terms are given in \citep{}, ...

%{\color{red} [it might be difficult to find exact references for all updates but they are all rather simple]}

%###########################################
\section{Bayesian Experimental Design}\label{sec:Bayes_ed}
%###########################################

The task is to find the feedback that maximises the expected information gain:
\begin{equation}
  j^* = \argmax_{j\notin \mathcal{F}} \E_{p(\tilde{f}_j|\bD)}\left[ \sum_i \KL{p(\tilde{y}|\bD, \bm{x}_i, \tilde{f}_j)}{p(\tilde{y}|\bD, \bm{x}_i)}\right],
\end{equation}
where $\mathcal{F}$ is the set of feedbacks that have already been given (to simplify notation, those are here assumed included in $\bD$) and the summation over $i$ goes over the training dataset. The evaluation of the expected information gain is described in the following.

The posterior predictive distribution is approximated as Gaussian:
\begin{equation}\label{eq:posterior_predictive}
  p(\tilde{y} | \bD, \tilde{x}) \approx \normalpdf(\tilde{y} | \tilde{\bm{x}}\tp \bar{\bm{m}}, \tilde{\bm{x}}\tp \bar{\bm{\Sigma}} \tilde{\bm{x}} + \bar{s}^2),
\end{equation}
where $\bar{s}^2 = \frac{\bar{\beta}_{\sigma}}{\bar{\alpha}_{\sigma}}$ is the posterior mean approximation for the residual variance.

Similarly, the posterior predictive distributions of the feedbacks for the two feedback types follow as approximate Gaussian and Bernoulli distributions:
\begin{align}
  p(\tilde{f}_{w,j} | \bD) &\approx \normalpdf(\tilde{f}_{w,j} | \bar{m}_j, \bar{\Sigma}_{jj} + \omega^2), \\
  p(\tilde{f}_{\gamma,j} | \bD) &\approx \bernoullipdf(\tilde{f}_{\gamma,j} | \pi \bar{\rho}_j + (1 - \pi) (1 - \bar{\rho}_j)).
\end{align}

The information gain between the predictive distributions is
\begin{equation}
  \KL{p(\tilde{y}|\bD, \tilde{\bm{x}}, \tilde{f}_j)}{p(\tilde{y}|\bD, \tilde{\bm{x}})} = \frac{1}{2}\left[\log \frac{\tilde{\bm{x}}\tp \bar{\bm{\Sigma}} \tilde{\bm{x}} + \bar{s}^2}{\tilde{\bm{x}}\tp \bar{\bm{\Sigma}}_{\tilde{f}} \tilde{\bm{x}} + \bar{s}_{\tilde{f}}^2} + \frac{\tilde{\bm{x}}\tp \bar{\bm{\Sigma}}_{\tilde{f}} \tilde{\bm{x}} + \bar{s}_{\tilde{f}}^2 + (\tilde{\bm{x}}\tp \bar{\bm{m}}_{\tilde{f}} - \tilde{\bm{x}}\tp \bar{\bm{m}} )^2}{\tilde{\bm{x}}\tp \bar{\bm{\Sigma}} \tilde{\bm{x}} + \bar{s}^2} - 1 \right].
\end{equation}

As running the EP algorithm to full convergence would be too costly for evaluating a large number of candidates, we approximate the posterior distribution with the new feedback with partial EP updates. This is similar to the approach of \citet{seeger2008bayesian} and ~\citet{hernandez2013generalized} for experimental design for sparse linear model. We consider the two types of feedback separately.

In the case of feedback directly on the regression weight, we add the corresponding site term (which is already of Gaussian form and does not need approximation, as noted above) and do not update the approximations of the other site terms (including assuming $\bar{s}_{\tilde{f}}^2 = \bar{s}^2$). The new posterior approximation of $\bw$ with these assumptions is
\begin{align}
  \bar{\bm{\Sigma}}_{\tilde{f}_{w,j}} &= (\bar{\bm{\Sigma}}^{-1} + T \bm{e} \bm{e}\tp)^{-1}, \\
  \bar{\bm{m}}_{\tilde{f}_{w,j}} &= \bar{\bm{\Sigma}}_{\tilde{f}_{w,j}} (\bar{\bm{\Sigma}}^{-1} \bar{\bm{m}} + h \bm{e})\label{eqn:new_mean},
\end{align}
where $\bm{e}$ is a vector of zeros except for $1$ at $j$th element, $T = \frac{1}{\omega^2}$, and $h = \frac{\tilde{f}_{w,j}}{\omega^2}$. Notably, $\bar{\bm{\Sigma}}^{-1}$ and $\bar{\bm{\Sigma}}^{-1} \bar{\bm{m}}$ are the precision and the precision-adjusted mean of the posterior approximation without the new feedback and are directly available from the previous EP approximation. The new posterior covariance is independent of the value of the feedback $\tilde{f}_{w,j}$ and it can be efficiently evaluated using the matrix inversion lemma as $\bar{\bm{\Sigma}}_{\tilde{f}} = \bar{\bm{\Sigma}} - \frac{1}{T^{-1} + \bar{\Sigma}_{jj}} \bar{\bm{\Sigma}} \bm{e} \bm{e}\tp \bar{\bm{\Sigma}}$. Furthermore, the expectation over the feedback in the expected information gain affects only the term with the squared difference of the means. This is
\begin{align}
  \E_{p(\tilde{f}_j|\bD)}\left[  (\tilde{\bm{x}}\tp \bar{\bm{m}}_{\tilde{f}} - \tilde{\bm{x}}\tp \bar{\bm{m}} )^2 \right] &= \E_{p(\tilde{f}_j|\bD)}\left[ \left(\frac{T_{jj}}{1 + T \bar{\Sigma}_{jj}} \tilde{\bm{x}}\tp \bar{\bm{\Sigma}} \bm{e}\right)^2 \left(\frac{h}{T} - \bar{m}_j\right)^2 \right] \\
  &= \left(\frac{T}{1 + T \bar{\Sigma}_{jj}} \tilde{\bm{x}}\tp \bar{\bm{\Sigma}} \bm{e}\right)^2 (\bar{\Sigma}_{jj} + \omega^2),
\end{align}
where the first equality follows from substituting the Equation \ref{eqn:new_mean} and using the matrix inversion lemma, and the second equality from $\frac{h}{T} = \tilde{f}_{w,j}$ and the remaining expectation being equal to the variance of the predictive distribution of the feedback.

In the case of relevance feedback, we add the corresponding site term for the feedback and run single EP update on it and the corresponding prior term $p(w_j|\gamma_j)$. These updates are purely scalar operations and do not require any costly matrix operations. Other site term approximations are not updated. The new posterior approximation of $\bw$ with these assumptions is
\begin{align}
  \bar{\bm{\Sigma}}_{\tilde{f}_{w,j}} &= (\bar{\bm{\Sigma}}^{-1} + T \bm{e} \bm{e}\tp)^{-1}, \\
  \bar{\bm{m}}_{\tilde{f}_{w,j}} &= \bar{\bm{\Sigma}}_{\tilde{f}_{w,j}} (\bar{\bm{\Sigma}}^{-1} \bar{\bm{m}} + h \bm{e}),
\end{align}
where $T = [\bar{\bm{\Sigma}}_{\tilde{f}_{\gamma,j}}^{-1}]_{jj} - [\bar{\bm{\Sigma}}^{-1}]_{jj}$ and $h = [\bar{\bm{\Sigma}}_{\tilde{f}_{\gamma,j}}^{-1} \bm{m}_{\tilde{f}_{\gamma,j}}]_j - [\bar{\bm{\Sigma}}^{-1} \bar{m}]_j$. That is, now $T$ and $h$ are the changes in the precision and the precision adjusted mean in the $j$th feature and these are available with cheap scalar operations. The expectation over the value of the feedback in the expected information gain is in this case a sum of two terms and we evaluate both of the terms separately using the above scheme. Again, we use the matrix inversion lemma to avoid full inversions in computing the new posterior covariance.

%###########################################
\section{Additional Experiments}\label{sec:additional_plots}
%###########################################

\subsection{Synthetic data}

For the synthetic experiments with simulated data, we continue the study of the behaviour of our algorithm, through additional experiments and visualisations. The setting stays the same as in Sect.~\ref{sec:heatmap_dimension}, except for the specifications below.

\subsubsection{Heatmaps with varying number of training data}\label{sec:heatmap_traindata}

%For the setting described in Sect.~\ref{sec:heatmap_dimension}, 

We now study the performance when the number of training data varies from 1 to 50 (since we consider in particular small-samples settings). The dimensionality is fixed to 100, and the number of relevant features is 10.

\begin{figure}[ht]
\centering
    \subfigure[Feedback on coefficients' values]{\label{fig:sim_study_all_weights}
    \includegraphics[width=0.45\textwidth,keepaspectratio]{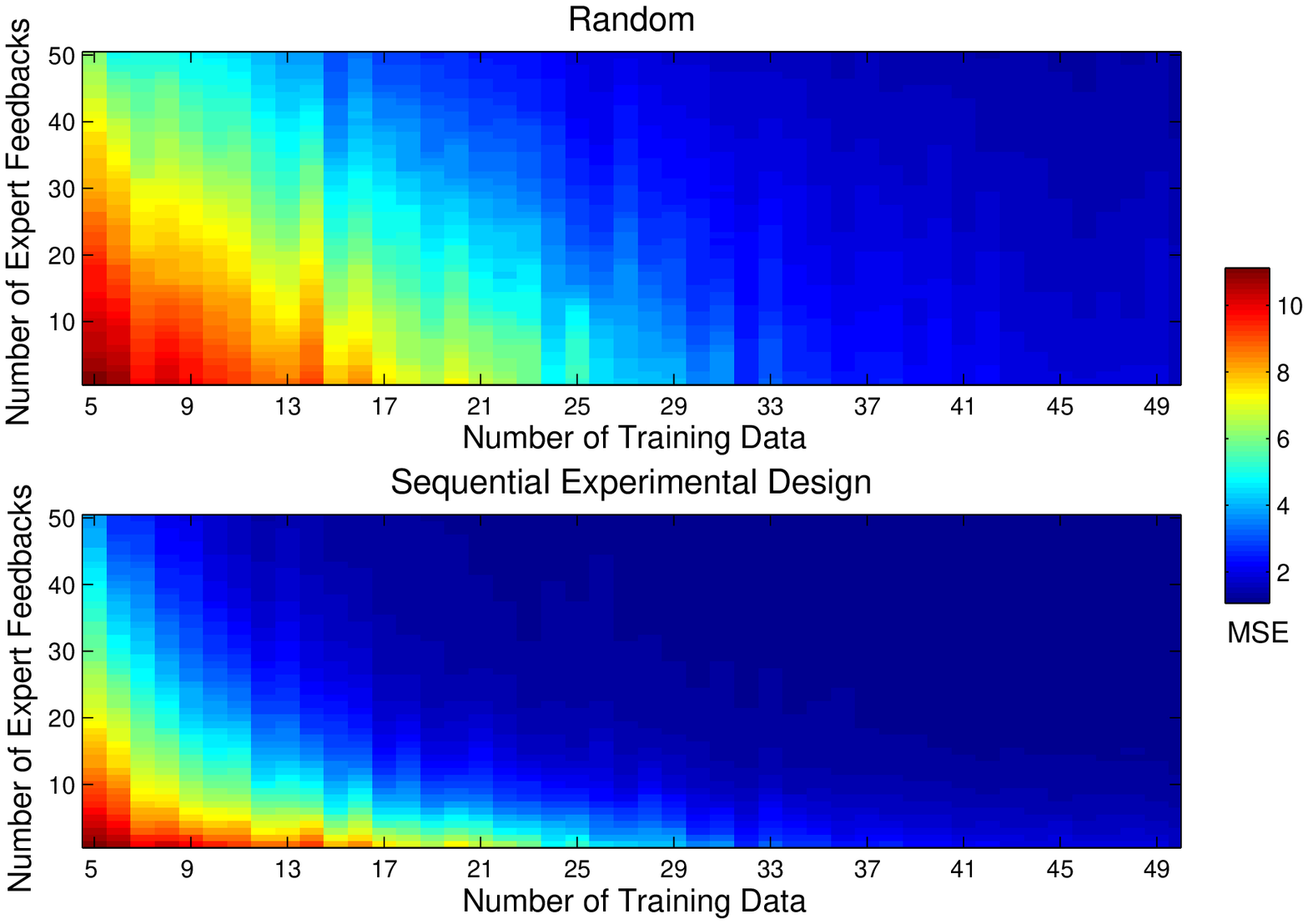}
    }
    \subfigure[Feedback on coefficients' relevances]{\label{fig:sim_study_all_relevance}
    \includegraphics[width=0.45\textwidth,keepaspectratio]{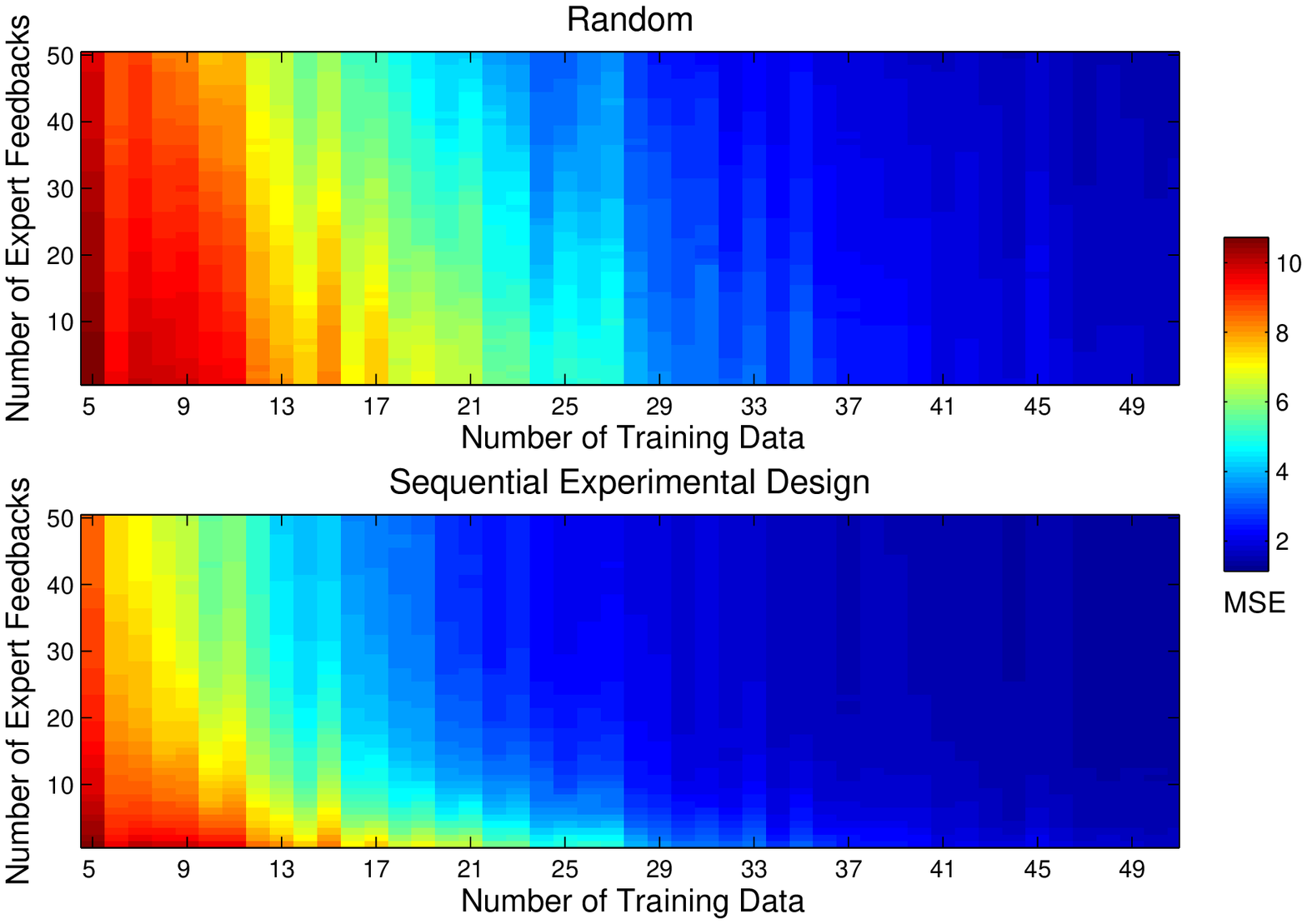}
    }

\caption{Mean squared errors with increasing number of training data. The number of relevant coefficients $m^*=10$ and the number of dimensions $m=100$. The MSE values are averages over 100 independent runs.}
\label{fig:heatmap_train_data}
\end{figure}

Fig.~\ref{fig:heatmap_train_data} illustrates the behaviour of our strategy and that of the random feature selection, for the previously described synthetic data setting with a fixed dimension $m=100$ and with increasing numbers of training data points $n=5,\dots,50$. For very small sample sizes ($n<10$), a difference between the performance of the two methods starts being visible after 20-30 received feedbacks. Then, for larger training samples sizes ($10<n<30$), the MSE reduction in our method is more visible from the first feedbacks, while for $n>30$, both strategies have a much smaller MSE. % (and the distinction between the two is no longer observable, due to the reduced colour scale in the heatmap). 

\subsubsection{Sequential vs Non-sequential Experimental Design}\label{sec:seq_vs_nonseq}

\begin{figure}[ht]
\centering
   \subfigure[Feedback on coefficients' values]{\label{fig:sim_study_weight_100_10}
       \includegraphics[width=0.45\columnwidth,keepaspectratio]{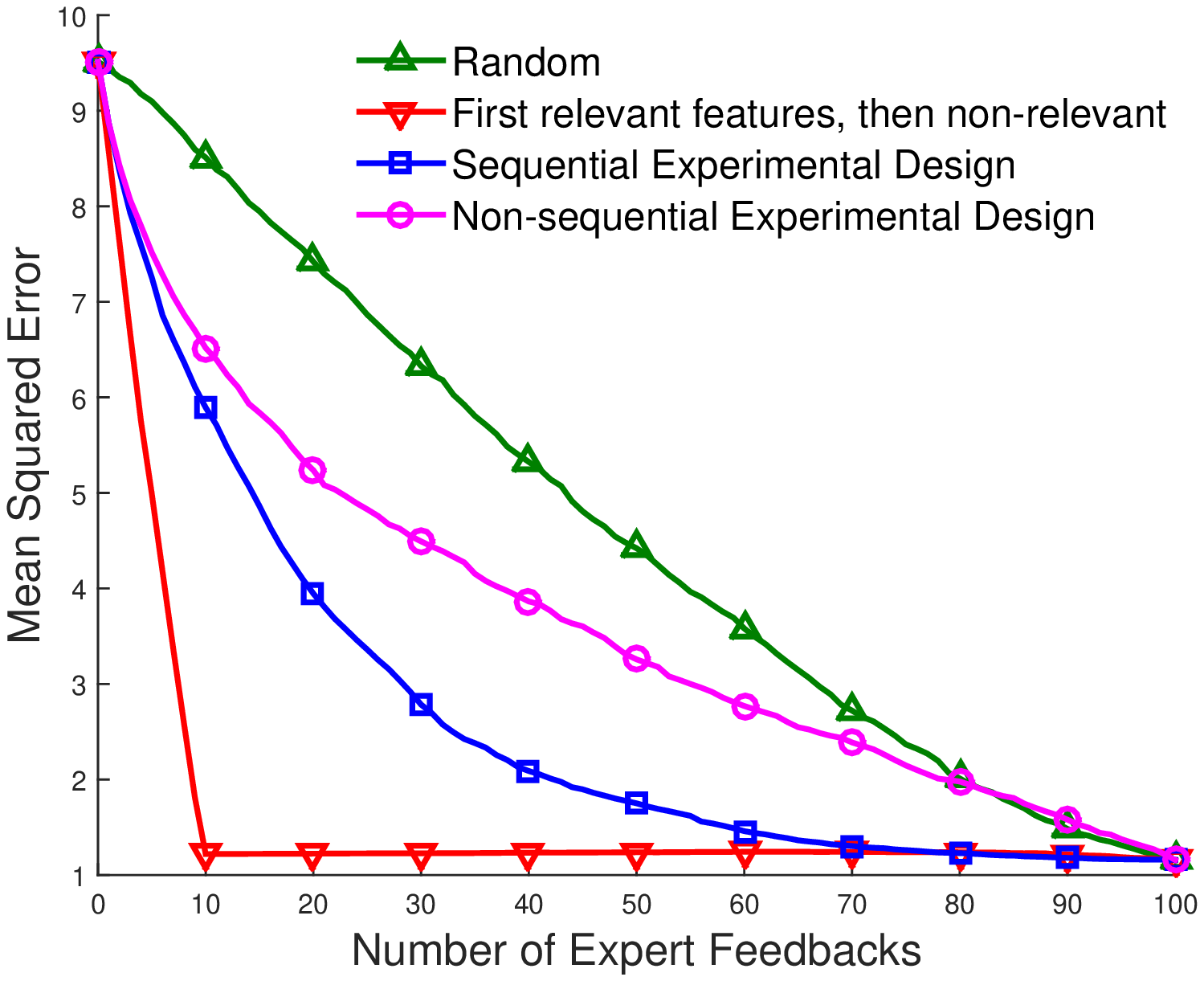}
   }
   \subfigure[Feedback on coefficients' relevances]{\label{fig:sim_study_relevance_100_10}
       \includegraphics[width=0.45\columnwidth,keepaspectratio]{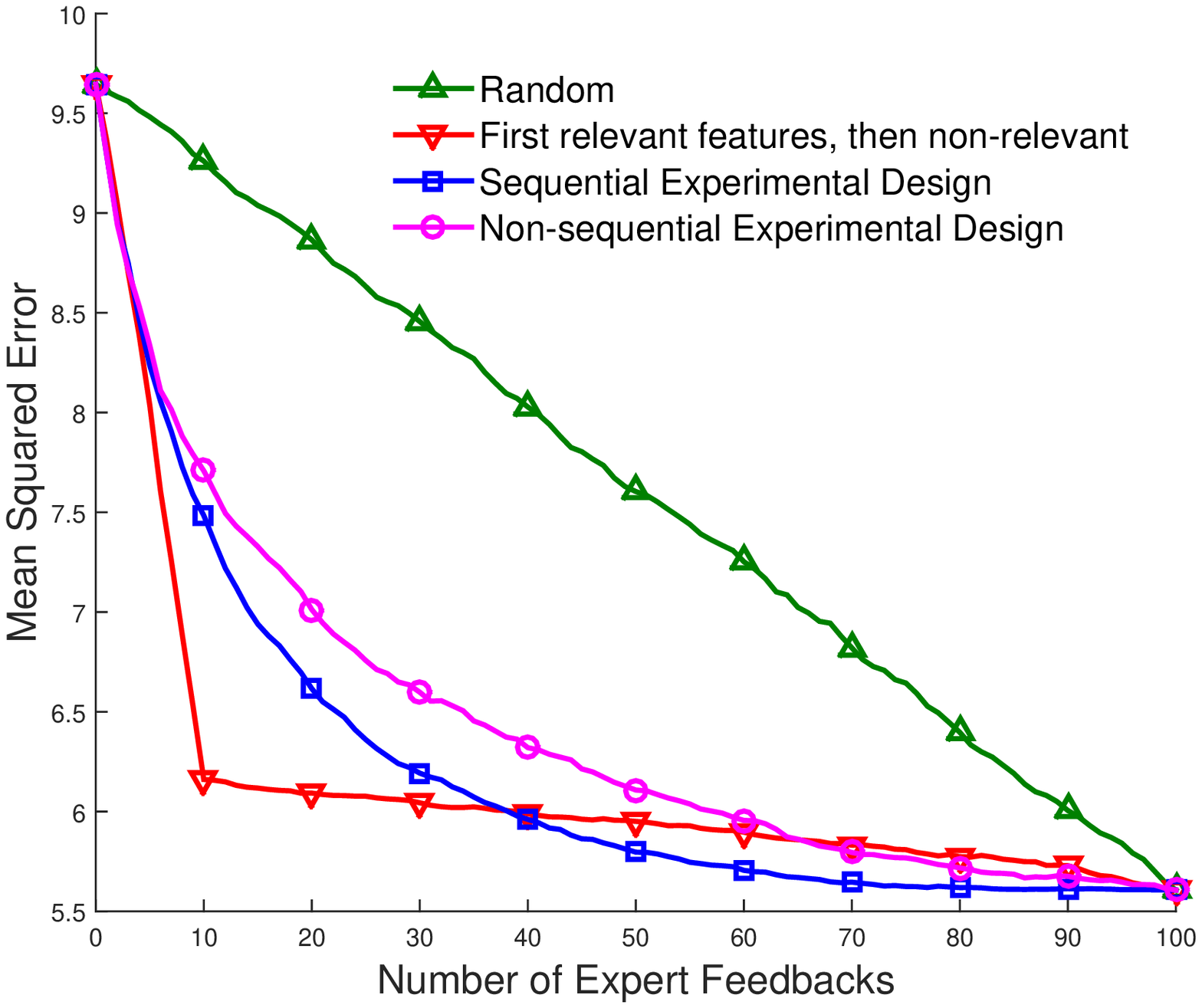}
   }
\caption{MSE for all query algorithms, with simulated data, for feedback on coefficient values %Fig.~\ref{fig:sim_study_weight_100_10}
and relevance. Note that the red strategy is not available in practice.} %Fig.~\ref{fig:sim_study_relevance_100_10}
\label{fig:sim_study_100_10_bothmodels}
\end{figure}

For a simple setting with simulated data, we now study the difference between our method and its non-sequential version for the two feedback models discussed previously: user feedback on the coefficients and on their relevance. The non-sequential version chooses the sequence of features to be queried before observing any expert feedback. We note that the behaviour and ranking of the query algorithms remain similar to the one observed in the previous plots. 
In Fig.~\ref{fig:sim_study_100_10_bothmodels}, we consider a ``small $n$, large $p$'' scenario, with $n=10, m=100, m^*=10$ and we report the average MSE value over 500 runs.

\begin{figure}[ht]
\centering
   \subfigure[MSE on training data]{\label{fig:sim_study_weight_100_10_tr}
       \includegraphics[width=0.45\columnwidth,keepaspectratio]{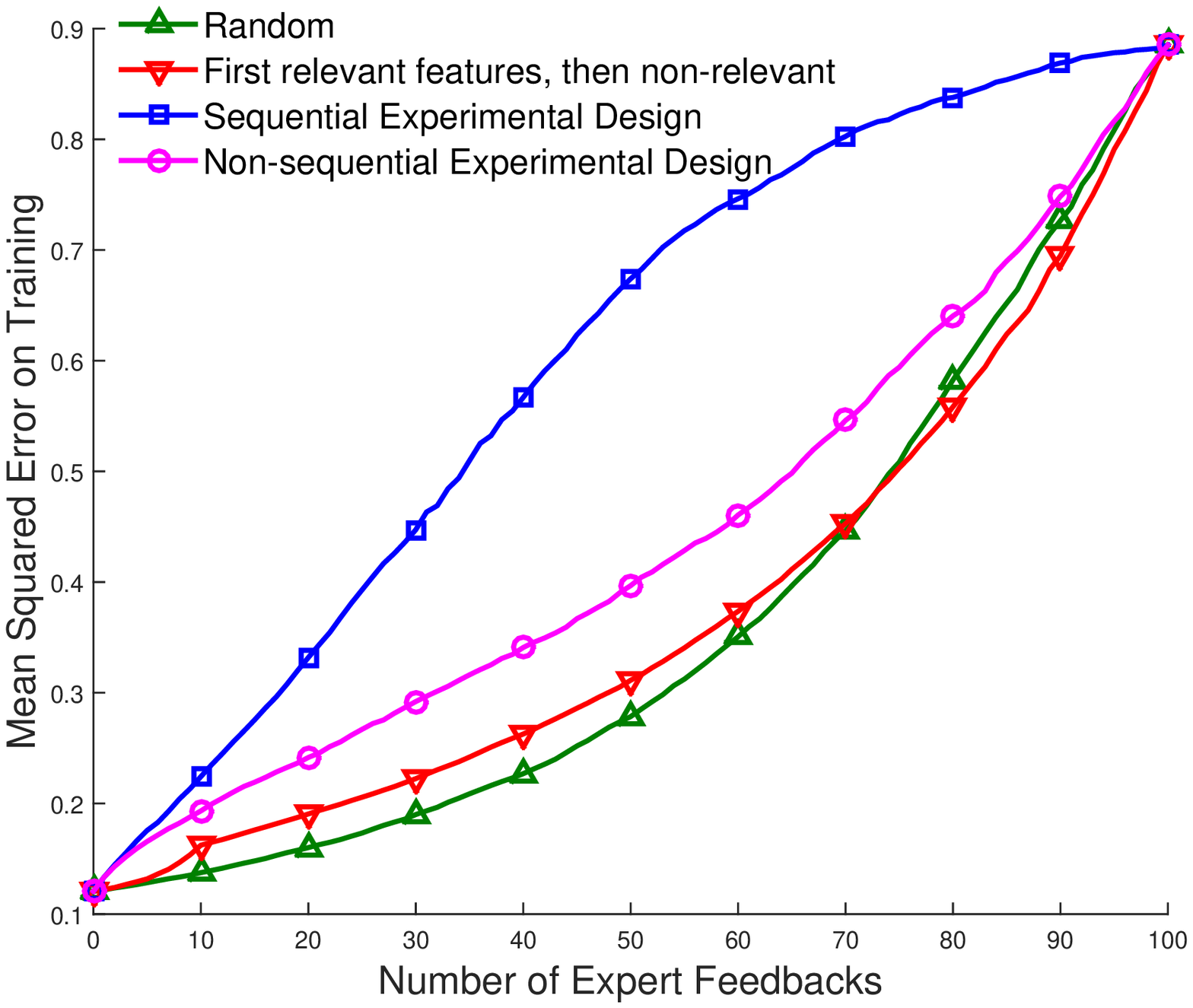}
   }
   \subfigure[Accumulated average suggestion]{\label{fig:sim_study_weight_100_10_suggestions}
       \includegraphics[width=0.45\columnwidth,keepaspectratio]{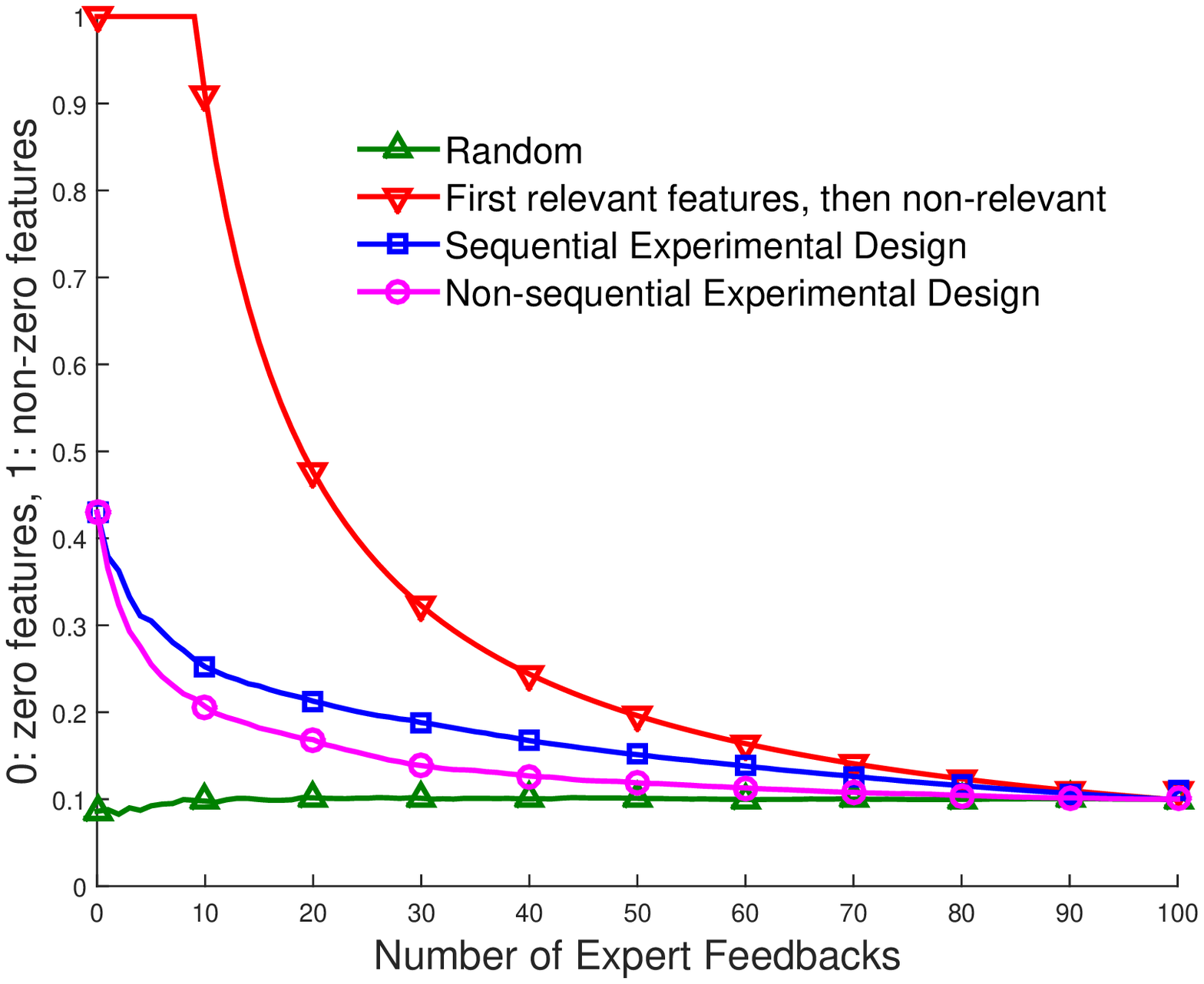}
   }
\caption{MSE on the training data and accumulated average suggestion behavior for all query algorithms, with simulated data, for the case where feedback is on coefficient values.} 
\label{fig:sim_study_100_10_bothmodels_tr}
\end{figure}

%\textbf{Results.} 
%
The results in the plots are shown for an increasing number of feedbacks, that gets to the number of dimensions, when all methods converge. However, if we consider the plausible scenario when the number of user interactions are limited, one can notice that compared to the other methods, the performance of both experimental design methods have a more important decrease in prediction loss even in the first iterations. 

This reflects the fact that both experimental design strategies manage to identify and ask with priority about the most informative coefficients. This is more evident in the feedback model about coefficient relevance (Fig.~\ref{fig:sim_study_relevance_100_10}), where the performance of the two experimental design strategies is very close to the strategy that first suggests only relevant features. However, one can also notice an improved performance for the sequential version of the experimental design strategy. Indeed, the more carefully selected sequence of queries done by the sequential experimental design strategy, manages to reduce the prediction error faster, compared to the non-sequential selection, where the observed expert feedback is not taken into account. Also, as expected, the difference between the sequential and non-sequential experimental designs is more significant in the case of the stronger feedback model  on coefficients values (Fig.~\ref{fig:sim_study_weight_100_10}).

\subsection{Comparison of Training and Test Set Errors and the Average Accumulated Suggestion Behaviour}\label{sec:trainingerror}

We can get some insight into the behaviour of the approach by comparing the training and test set errors shown in Fig.~\ref{fig:sim_study_weight_100_10} and Fig.~\ref{fig:sim_study_weight_100_10_tr} for the simulated data scenario described in the previous section with feedback on the coefficient values. The training set error begins to increase as a function of the number of expert feedbacks. This happens because the model without any feedbacks has exhausted the information in the training data (to the extent allowed by the regularizing priors) and fits the training data well. The user feedback, however, moves the model away from the training data optimum and towards better generalization performance. Indeed, the MSE curves for the training and test errors converge close to each other as the number of feedbacks increases. Moreover, the convergence is faster for the query algorithms that start by suggesting the features with non-zero effects implying that they are more informative (Fig.~\ref{fig:sim_study_weight_100_10_suggestions}).

%*********************************AMAZON TABLE**********************************
\subsection{Expert Knowledge Elicitation vs.\ Collecting More Samples}\label{sec:amazon_info_gain_comparison}

\begin{comment}
% this is the table in the submitted version
\begin{table}[h]
\centering
%\resizebox{0.6\textwidth}{!}{  
    \begin{tabular}{c|c|c|c|c|c}
     & \multicolumn{2}{c|}{More Samples}  & \multicolumn{2}{c|}{More Feedback} \\
    \hline
   \textsc{MSE} & \textsc{Random} & \textsc{Active}~\cite{seeger2008bayesian} & \textsc{Random} & \textsc{SeqExpDes}  \\
    \hline
    2  & 5 & 4  & 50 & 5
    \\
    1.975 & 20 & 10 & 122 & 18
    \\
    1.95 & 36 & 43 & $>$200 & 33
    \\
    1.925 & 52 & 68 & $>$200 & 51
    \\
    l.9 & 68  & 95 & $>$200 & 70 
    \\
    1.875  & 116 & 142 & $>$200 & 105 
    \end{tabular}
%}
\caption{Number of samples/feedbacks needed to reach a particular MSE level in Amazon dataset. The results are averages over 100 independent runs.}
\label{table:realdata_simuser}
\end{table}
\end{comment}

% this is the table based on the new set of runs
\begin{table}[h]
\centering
%\resizebox{0.6\textwidth}{!}{  
    \begin{tabular}{c|c|c|c|c|c}
     & \multicolumn{2}{c|}{More Samples}  & \multicolumn{2}{c|}{More Feedback} \\
    \hline
   \textsc{MSE} & \textsc{Random} & \textsc{Active}~\cite{seeger2008bayesian} & \textsc{Random} & \textsc{SeqExpDes}  \\
    \hline
    2.025  & 8 & 4  & 73 & 9
    \\    
    2  & 15 & 7  & 152 & 19
    \\
    1.975 & 29 & 12 & $>$200 & 31
    \\
    1.95 & 44 & 43 & $>$200 & 43
    \\
    1.925 & 59 & 71 & $>$200 & 64
    \\
    l.9 & 98  & 92 & $>$200 & 95 
    \\
    1.875  & $>$200 & 144 & $>$200 & 136 
    \end{tabular}
%}
\caption{Number of samples/feedbacks needed to reach a particular MSE level in Amazon dataset. The results are averages over 100 independent runs.}
\label{table:realdata_simuser}
\end{table}

\subsection{User Study}\label{sec:user_study_supplementary}

%\todom{Should someting about the t test be reported here?}

We complement the analysis of the results of the user study with two illustrations. 

\begin{figure}[h]
\centering
   \subfigure[Percentage of improvement in MSE]{\label{fig:user_study_percentage}
    \includegraphics[width=0.45\textwidth,keepaspectratio]{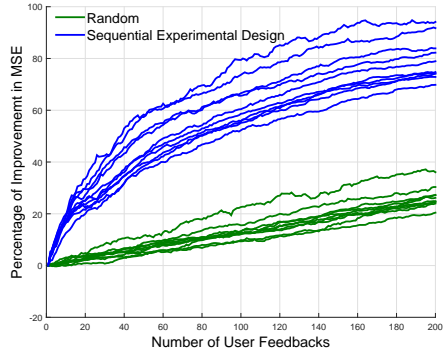}
    }
   \subfigure[Accumulated average suggestion]{\label{fig:user_study_suggestions}
    \includegraphics[width=0.45\textwidth,keepaspectratio]{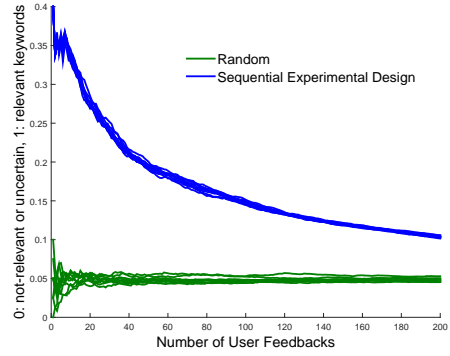}
    }
\caption{User study results: MSE for 10 participants, Amazon data.}
\label{fig:user_study_supplementary}
\end{figure}

First, to compare the convergence speed of different methods, we normalised the MSE improvements at each iteration by the amount of total improvement obtained by each of the users, when considering all their individual feedback. Figure \ref{fig:user_study_percentage} depicts the convergence speed of methods based on this measure. As can be seen from the figure, for all participants, the proposed method was able to capture most of the participants's knowledge with small budget of feedback queries  (stabilizing at around 200 out of the total 824 features in the considered subset of Amazon data). 

\begin{figure}[h]
\hspace{-1.8cm}
    \includegraphics[width=1.2\textwidth,keepaspectratio]{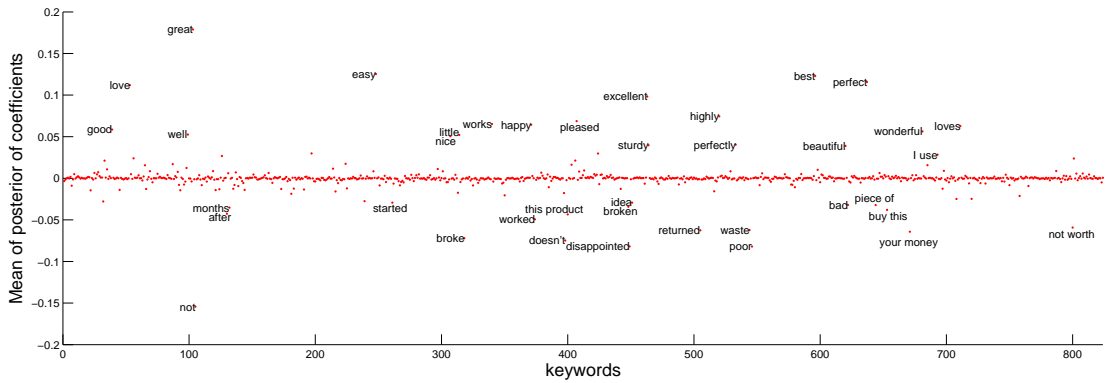}
\caption{Mean of the coefficients for all words trained over all reviews in Amazon dataset. The name of the 39 relevant words is shown beside their coefficient values (shown in the Y axis).}
\label{fig:user_study_gt}
\end{figure}

Then, in Figure \ref{fig:user_study_suggestions}, we show the average percentage of relevant words that were asked from the participants at each iteration. It is evident from the figure that the proposed algorithm started by mostly asking about limited relevant words. The relevant words were identified by considering all the data in Amazon dataset and training an spike and slab model and then choosing words with $\mathbf{E}[\gamma_j]>0.7$ (words with high posterior inclusion probability). Based on this threshold, only 39 words from 824 words were considered as relevant. The coefficients of all words in the spike-and-slab model, along with the names of relevant words are shown in Fig.~\ref{fig:user_study_gt}.

%\newpage

%\putbib
%\end{bibunit}

\end{document}